\documentclass[letterpaper]{article} 
\usepackage{aaai23}  
\usepackage{times}  
\usepackage{helvet}  
\usepackage{courier}  
\usepackage[hyphens]{url}  
\usepackage{graphicx} 
\usepackage{natbib}  
\usepackage{caption} 
\usepackage{algorithm}
\usepackage{algorithmicx}
\usepackage{algpseudocode}

\usepackage{url}
\usepackage{amsfonts}
\usepackage{graphicx}
\usepackage{bm}
\usepackage{xcolor}
\usepackage{amsmath}
\usepackage{xcolor}         

\usepackage{algorithm}
\usepackage{tabularx}
 \usepackage{verbatim}
\usepackage{subcaption}
\usepackage{times}
\usepackage{epsfig}
\usepackage{graphicx}
\usepackage{booktabs}
\usepackage{multirow}
\usepackage{multicol}
\usepackage{bm}
\usepackage{colortbl}

\usepackage{microtype}
\usepackage{graphicx}
\usepackage{booktabs} 
\usepackage{booktabs}
\usepackage{graphics}
\usepackage{graphicx}
\usepackage{epstopdf}
\usepackage{caption}
 \usepackage{wrapfig}
\usepackage{enumitem}
\usepackage{psfrag}
\usepackage{graphicx}
\usepackage{array}
\usepackage{multirow}
\usepackage{bm}
\usepackage{verbatim}
\usepackage{relsize}
\usepackage{algpseudocode}
\usepackage{tabularx}
 \usepackage{verbatim}
\usepackage{bm}
\usepackage{lipsum}
\usepackage{tabularx}
\usepackage{mwe} 
\usepackage{url}            
\usepackage{booktabs}       
\usepackage{nicefrac}       
\usepackage{microtype}      
\usepackage{times}
\usepackage{epsfig}
\usepackage{graphicx}
\usepackage{mathrsfs} 
\usepackage{wrapfig}
\usepackage{lscape}
\usepackage{rotating}

\usepackage{wrapfig}

\usepackage{bm}

\usepackage{setspace}
\usepackage{xspace}

\usepackage{newfloat}
\usepackage{listings}
\DeclareCaptionStyle{ruled}{labelfont=normalfont,labelsep=colon,strut=off} 
\floatstyle{ruled}
\newfloat{listing}{tb}{lst}{}
\floatname{listing}{Listing}
\title{Overcoming Concept Shift in Domain-Aware Settings through Consolidated Internal Distributions}
\author{
    Mohammad Rostami and Aram Galstyan\\
}
\affiliations{
    Information Sciences Institute, University of Southern California
    \\
    \{mrostami, galstyan\}@isi.edu
}

\usepackage{bibentry}

\begin{document}

\maketitle

\begin{abstract}
We develop an algorithm to improve the   performance of a pre-trained model under \textit{concept shift} without retraining the model from scratch when only unannotated samples of initial concepts are accessible. We model this problem as a domain adaptation problem, where the source domain data is inaccessible during model adaptation. The core idea is based on consolidating the intermediate internal distribution, learned to represent the source domain data, after adapting the model. We provide theoretical analysis and conduct extensive experiments   to demonstrate that the proposed method is effective. 
\end{abstract}

\section{Introduction}

When a trained neural network is used for prediction, the assumption is that the testing samples are drawn from the training distribution. However, concept shift is a natural phenomenon when a model is used in different domains~\cite{vorburger2006entropy}.  Similar to other machine learning techniques, deep learning is vulnerable with respect to these distributional shifts  during the model execution time. Distributional shifts would lead to mode performance degradation which makes model retraining, i.e, \textit{model adaptation}, inevitable in order to make the model generalizable again in new domains. 
Adapting deep neural networks is conditioned on availability of massive labeled datasets which may not  always be feasible due to prohibitive costs of manual data annotation. 
  Finetuning  is a common strategy to improve model generalization in the presence of domain shift. Despite being effective in reducing the effect of domain shift,  finetuning still requires annotated data in the target domain.

\cite{mirza2022norm,wang2022cross}

Unsupervised domain adaptation (UDA) is a similar learning setting in which
only \textit{unlabeled data} is accessible in the target domain~\cite{glorot2011a} along with labeled data in the source domain.
The goal in UDA is to leverage from the source domain  data to adapt the deep neural network to generalize well in the target domain. 
 An effective approach for UDA is to align distributions of both domains by mapping data into   a latent domain-invariant space~\cite{daume2009frustratingly}. 
  As a result, a classifier that is trained using  the source labeled data features in this space will generalize well in the target domain. UDA methods model this latent space as the output of a shared deep encoder. The encoder network is trained such that the source and the target domains share a similar   distribution in its output. This training procedure has been implemented using either  adversarial learning~\cite{he2016deep,sankaranarayanan2018generate,pei2018multi,zhang2019domain,long2018conditional} or by directly minimizing the distance between the two distributions in the embedding~\cite{long2015learning,ganin2014unsupervised,long2017deep,kang2019contrastive,rostami2019deep}. 

 Most existing unsupervised domain adaptation (UDA) algorithms consider a joint learning setting, where the model is trained   jointly on both the target domain unlabeled data and the source domain labeled data. As a result, these algorithms cannot be used for sequential model adaptation. Note that although  a few source-free domain adaptation algorithms have been developed recently~\cite{kundu2020universal,li2020model,liu2021source}, these methods use adversarial learning to memorize the source domain to generate source domain pseudo-data points for model retraining which necessitates using additional networks, rather directly adapting the base classifier network, which makes the model and the optimization problem more complex . Our goal is to adapt the classifier model   to generalize well in the target domain using solely the target domain unlabeled data to tackle domain shift. 
 This setting can be considered as an improvement over using an off-the-shelf pre-trained model when unlabeled data in a source domain is available. Our approach also relaxes  the necessity of sharing training data between the two domains.
 
 
{\em \bf Contributions:} we develop a sequential model adaptation algorithm which is based on learning a parametric internal distribution for the source domain   distribution in a shared embedding space. This internally learned distribution    is used to align the source and the target distributions. 
In order to adapt the model to work well the target domain, we draw samples from the estimated internal distribution and enforce the target domain to share the same    distribution in the embedding space by minimizing the distance between the two distributions.  
We   provide a theoretical justification for the proposed algorithm, by establishing an upperbound for the expected risk in the target domain, and demonstrating that our algorithm minimizes this upperbound. We conduct   experiments on five   benchmarks and observe that our algorithm compares favorably  against  SOTA   UDA methods.

\section{Background and Related Work}     
 UDA is closely related to our learning setting. Several discrepancy measures have been used in the literature to align two distributions to address UDA. A group of methods match the first-order and the second order statistics of the source and the target domains. This includes methods that use  the Maximum Mean Discrepancy (MMD)~\cite{long2015learning,long2017deep} and correlation alignment~\cite{sun2016deep}.
 A more effective approach is to use a probability distance metric that captures distributional differences in higher order statistics. The Wasserstein distance (WD)~\cite{courty2017optimal,damodaran2018deepjdot} is such an example that is also a suitable metric for deep learning   due to having non-vanishing gradients.  This property is helpful because deep learning optimization problems are usually solved using  the first-order   optimization methods that rely on the objective function non-vanishing gradients.
 Damodaran et al.~\cite{damodaran2018deepjdot} used the WD for domain alignment in a UDA setting which led to considerable performance improvement compared to the methods that rely on matching only lower-order statistical moments~\cite{long2015learning,sun2016deep}.
We   rely on  the sliced Wasserstein distance (SWD) variant of WD~\cite{kolouri2018sliced} for domain alignment which   can be computed more efficiently due to its closed-form formulation and has been used for UDA successfully~\cite{lee2019sliced,stan2021unsupervised}.
More recently, secondary mechanism have been used to improve upon the primary alignment mechanism~\cite{mirza2022norm,wang2022cross}.
To mitigate  negative transfer, some methods   separate between transferable and domain-specific knowledge
\cite{dong2020can,dong2021and}.

Most existing UDA methods use a strong assumption. It is assumed that the source and the target domain data are accessible simultaneously and the model is trained jointly on both datasets. Sequential model adaptation can be considered as a more challenging setting, where the source domain data is inaccessible after an initial model  training  for various reasons, including privacy and security concerns.
It is different from continual learning~\cite{chen2018lifelong,rostami2020using} in that not all tasks are labelled and addressing forgetting effects~\cite{kirkpatrick2017overcoming,rostami2020generative} is the primary concern.
 By addressing the sequential model adaptation setting, we can also address UDA when the source domain data cannot be shared~\cite{rostami2018multi}. This learning  setting for domain adaptation has been explored for non-deep models~\cite{dredze2008online,jain2011online,wu2016online}. However, these works address sequential model adaptation when the input distribution can be estimated with a parametric distribution and the base models have a small number of parameters. Hence, it is not trivial to extend the above works  for the end-to-end training procedure  of deep neural networks when deep learning is necessary for decent performance. Recently, adversarial learning has been used to address source-free DA~\cite{kundu2020universal,li2020model}. These methods memorize the source domain using additional networks to generate a source pseudo-dataset that can be used as a surrogate for the   source dataset. While these methods relax the need for  access to the source data, generating realistic data points for complex datasets is infeasible. Moreover, having an additional network makes optimization problem   challenging. 
Quite differently, we rely  an internally learned distribution by a base model for the source domain to align the two source and target domain distributions indirectly in an embedding space to adapt the base classifier   sequentially without using additional networks. We will improve upon a previous work that does not assume a source-free constraint~\cite{rostami2019deep}.

When a deep network is trained on a classification problem,  the model would have decent performance only if  the data points that belong to each  class  form a single cluster in an embedding space which is modeled by the network responses at higher layers. In other words, the source domain input distribution is transformed into a multimodal internal distribution, where each mode of the distribution 
encodes one of the classes. This internal distribution  encodes the knowledge gained from supervised learning in the source domain. Domain shift occurs when changes in the input distribution lead to discrepancies between the transformed input distribution and the internally learned distribution. UDA can be addressed by aligning the target domain distribution   with the learned source domain internal distribution. 
We     estimate the internal distribution  in the embedding   via a parametric Gaussian mixture model (GMM). Previously, this idea has been used to address learning settings different from our setting \cite{oliveira2019gmm,wu2005tracking,pfulb2021overcoming}
We develop an algorithm for 
addressing sequential model adaptation by  enforcing the target domain to share the same internal distribution.

\begin{figure} 
    \centering  
    \centering
    \includegraphics[width=\linewidth]{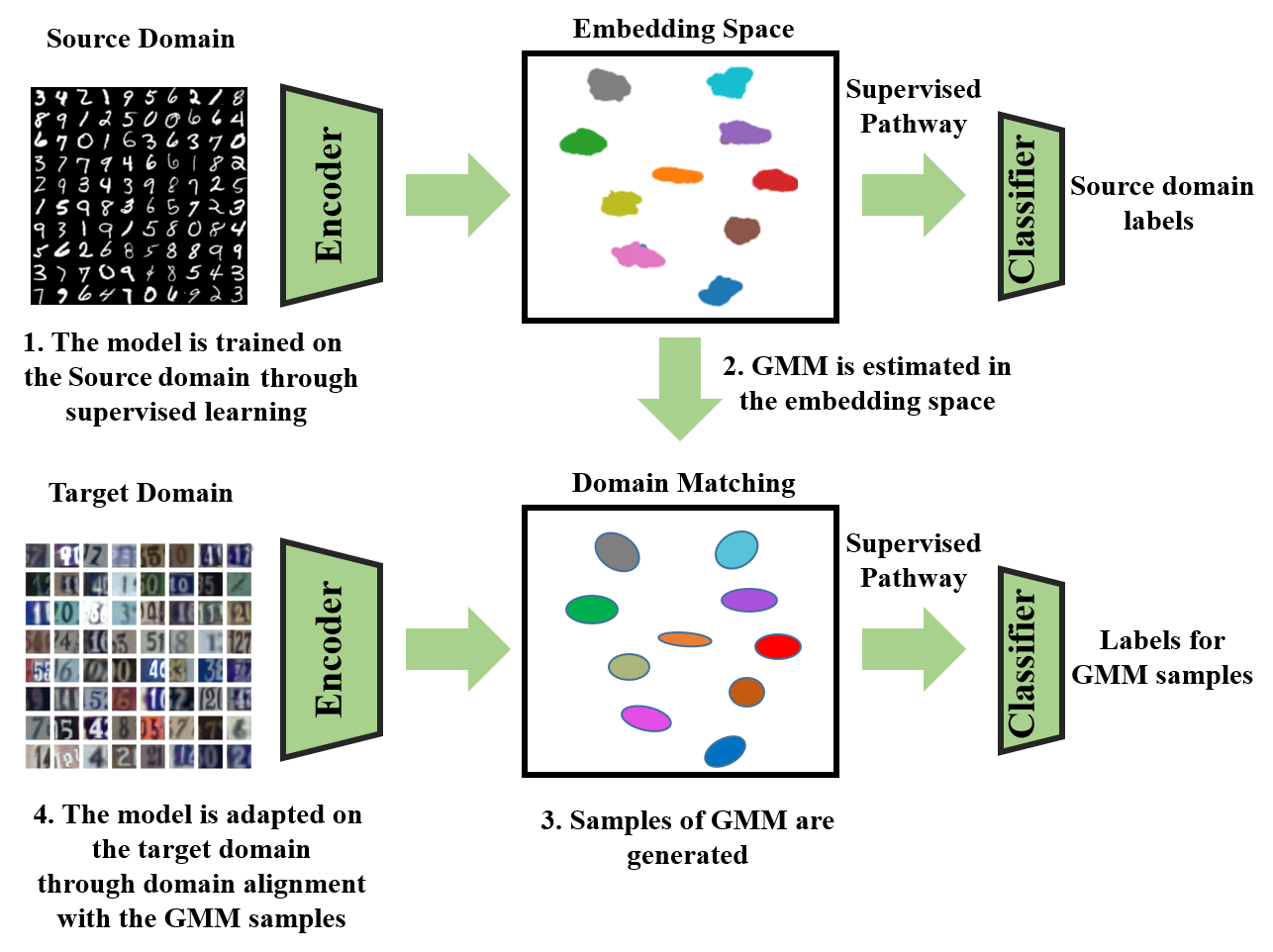}
         \caption{\small Architecture of the proposed   method: (top) a neural network is trained on a source domain with annotated data. The internally learned distribution at the output space of the encoder   is modeled via a GMM. (bottom) the model is updated to address concept shift in the target domain with annotated data by matching the target domain distribution with the GMM using a pseudo-dataset. }
         \label{figDL:LL}
         \vspace{-5mm}
\end{figure}

\section{Problem Statement}
Consider a source domain $\mathcal{S}$ which consists of the distribution $p_{S}(\bm{x})$ and the labeling function $f(\cdot):\mathbb{R}^d\rightarrow \mathcal{Y}\subset \mathbb{R}^k$.  Given  a   family of parametric functions $f_{\theta}$, e.g., a deep neural network with learnable parameter $\theta$, our goal is to solve for an optimal model with minimum expected risk, i.e., $\mathbb{E}_{\bm{x}^s\sim p_{S}(\bm{x})}(\mathcal{L}(f(\bm{x}^s),f_{\theta}(\bm{x}^s))$,  where $\mathcal{L}(\cdot)$ is a proper loss function. To this end,
we are given a labeled source dataset $\mathcal{D}_{\mathcal{S}}=(\bm{X}_{\mathcal{S}}, \bm{Y}_{\mathcal{S}})$,  with   $\bm{X}_{\mathcal{S}}=[\bm{x}_1^s,\ldots,\bm{x}_N^s]\in\mathcal{X}\subset\mathbb{R}^{d\times N}$   and   $\bm{Y}_{\mathcal{S}}=[\bm{y}^s_1,...,\bm{y}^s_N]\in \mathcal{Y}\subset\mathbb{R}^{k\times N}$, where  the data points are drawn i.i.d.   $\bm{x}_i^s\sim p_{S}(\bm{x})$. 
Given a sufficiently large dataset, we can solve for the optimal parameters  using the standard empirical risk minimization (ERM): $\hat{ \theta}=\arg\min_{\theta}\hat{e}_{\theta}(\bm{X}_{\mathcal{S}},\bm{Y}_{\mathcal{S}},\mathcal{L})=\arg\min_{\theta}\sum_i \mathcal{L}(f_{\theta}(\bm{x}_i^s),\bm{y}_i^s)$.  Consider that after training on the source domain, we   encounter a target domain sequentially with  an unlabeled dataset $\mathcal{D}_{\mathcal{T}}=(\bm{X}_{\mathcal{T}})$, where $\bm{X}_{\mathcal{T}}=[\bm{x}_1^t,\ldots,\bm{x}_M^t]\in\mathbb{R}^{d\times M}$ and $\bm{x}_i^t\sim  p_{T}(\bm{x})$. As a result, using ERM  is not feasible in the target domain. We know a priori that the two domains share the same classes but also distributional discrepancy exists between the   domains, i.e., $p_{S}\neq p_{T}$. This leads to poor generalization of $f_{\hat{\theta}}$ on the target domain due to domain shift.  Our goal is to adapt the model using solely the dataset $\mathcal{D}_{\mathcal{T}}$ such that it generalizes well in the target domain in the absence of source   dataset (see Figure~\ref{figDL:LL}).  

In order to circumvent the challenge of distributional gap between the two domains, we can decompose the  mapping $f_\theta(\cdot)$ into a deep encoder $\phi_{\bm{v}}(\cdot): \mathcal{X}\rightarrow \mathcal{Z}\subset \mathbb{R}^p$ and a classifier subnetwork $h_{\bm{w}}(\cdot): \mathcal{Z}\rightarrow \mathcal{Y}$ such that $f_\theta = h_{\bm{w}}\circ \phi_{\bm{v}}$, where $\theta=(\bm{w},\bm{v})$. Here,  $\mathcal{Z}$ denotes an intermediate embedding space between the input space and the label space in which we assume that the   classes have become geometrically separable.  Given $\hat{\theta}$ and $\mathcal{D}_{\mathcal{T}}$, if we   adapt $\phi_{\bm{v}}$ such that  the discrepancy between the source and target distributions, i.e., the distance between $\phi(p_{\mathcal{S}}(\bm{x}^s))$ and $\phi(p_{\mathcal{T}}(\bm{x}^t))$, is minimized in the embedding space (making the embedding  domain agnostic), then  the classifier $h_{\bm{w}}$ will generalize well on the target domain, despite   being trained using only the source labeled data points. 
 Many prior classic UDA algorithms use this strategy but implement it by assuming that $\mathcal{D}_{\mathcal{S}}$ is always accessible. This makes computing the distance between the distributions $\phi(p_{\mathcal{S}}(\bm{x}^s))$ and $\phi(p_{\mathcal{T}}(\bm{x}^t))$ feasible and hence UDA reduces to selecting a proper probability   metric and then   solving a typical deep learning minimization problem~\cite{long2015learning,long2017deep,damodaran2018deepjdot,courty2017optimal,damodaran2018deepjdot}. The major challenge in the sequential model adaptation setting   is that as opposed to the classical UDA framework, the term $\phi(p_{\mathcal{S}}(\bm{x}^s))$ cannot be  computed   and we need a new approach to estimate this term using a suitable surrogate term.

\section{Proposed Algorithmic Solution}
\label{sec:UDAproposdframework}
We propose to solve the challenge of sequential model adaptation through aligning the source and the target distribution indirectly via an intermediate internally learned distribution in the embedding space. We set a  softmax function at the output of the encoder just before passing data representations into the classifier subnetwork. As a result,
the classifier can be assumed as  a maximum \textit{a posteriori} (MAP) estimator which assigns a membership probability distribution to any given data point. This assumption implies that when   an optimal model is trained for the source domain,  the encoder would transform the source distribution in the input space into a multi-modal distribution $p_J(\bm{z})$  with $k$ components in the embedding space, each corresponding to one class (see Figure~\ref{figDL:LL}, right). This internal distribution emerges because the classes  should become separable in the embedding space as the result of learning for a generalizable model with a softmax layer. This internal distribution is a multimodal distribution in which each distribution mode would capture one of the classes. Performance degradation occurs when shifts in the input distribution violate boundaries between the classes in the embedding space.
 If we   update the model such that the internal distribution remains stable in the target domain
 after model adaptation, i.e., the source and the target domain would share similar internal distributions in the embedding space, then the classifier subnetwork would generalize well in the target domain due to negligible domain gap.

The empirical version of the internal source   distribution is encoded  by the source data samples $\{(\phi_{\bm{v}}(\bm{x}_i^{s}),\bm{y}_i^{s})\}_{i=1}^{N}$. We   consider that $p_J(\bm{z})$ is a GMM   with $k$ components: 
\begin{equation}
\small
p_J(\bm{z})=\sum_{j=1}^k \alpha_j
\mathcal{N}(\bm{z}|\bm{\mu}_j,\bm{\Sigma}_j),
\end{equation}  
where $\alpha_j$ denote mixture weights, i.e., prior probability for each class, $\bm{\mu}_j$ and $\bm{\Sigma}_j$ denote the mean and co-variance for each component.
Since we have labeled data points, we can compute  the GMM parameters using MAP estimates~\cite{rostami2021lifelong}. Let $\bm{S}_j$ denote the support set for class $j$ in the training dataset, i.e., $\bm{S}_j=\{(\bm{x}_i^s,\bm{y}_i^s)\in \mathcal{D}_{\mathcal{S}}|\arg\max\bm{y}_i^s=j \}$. Then, the MAP estimate for the GMM parameters is:  
\begin{equation}
\small
\begin{split}
&\hat{\alpha}_j = \frac{|\bm{S}_j|}{N},\hspace{2mm}\hat{\bm{\mu}}_j = \sum_{(\bm{x}_i^s,\bm{y}_i^s)\in \bm{S}_j}\frac{1}{|\bm{S}_j|}\phi_v(\bm{x}_i^s),\\& \hat{\bm{\Sigma}}_j =\sum_{(\bm{x}_i^s,\bm{y}_i^s)\in \bm{S}_j}\frac{1}{|\bm{S}_j|}\big(\phi_v(\bm{x}_i^s)-\hat{\bm{\mu}}_j\big)^\top\big(\phi_v(\bm{x}_i^s)-\hat{\bm{\mu}}_j\big).
\end{split}
\label{eq:MAPest}
\end{equation}    
Our major idea is to use this internal distributional estimate to circumvent the major challenge of sequential model adaptation. In order to adapt the model to work well for the target domain, we update the model such that the encoder subnetwork matches the target domain distribution with the internal distribution for the source domain in the embedding space, making the embedding domain-invariant. To this end, we can draw random samples from the internal distributional estimate and generate a labeled pseudo-dataset: $\mathcal{\hat{D}}=(\textbf{Z}_{\mathcal{P}},\textbf{Y}_{\mathcal{P}})$, where      $\bm{Z}_{\mathcal{P}}=[\bm{z}_1^p,\ldots,\bm{z}_{N_p}^p]\in\mathbb{R}^{p\times N_p}$,   $\bm{Y}_{\mathcal{P}}=[\bm{y}^p_1,...,\bm{y}^p_{N_p}]\in \mathbb{R}^{k\times N_p}$, $\bm{z}_i^p\sim \hat{p}_J(\bm{z})$, and the labels are ascribed according to the classifier subnetwork prediction. To generate a clean pseudo-dataset, we also set a threshold $\tau$ and include only those generated samples for which the classifier prediction confidence is greater than $\tau$.  The sequential model adaptation problem then reduces to solving   the    alignment  problem as the following:
\begin{equation}
\small
\min_{\bm{v},\bm{w}} \sum_{i=1}^N \mathcal{L}\big(h_{\bm{w}}(\bm{z}_i^p),\bm{y}_i^p\big)+\lambda D\big(\phi_{\bm{v}}(p_{\mathcal{T}}(\bm{X}_{\mathcal{T}})),\hat{p}_{J}(\bm{Z}_{\mathcal{P}})\big),
\label{eq:mainPrMatch}
\end{equation}  
where $D(\cdot,\cdot)$ denotes a probability distribution metric to measure the distributional discrepancy, and $\lambda$ is a trade-off parameter between the two  terms (see Figure~\ref{figDL:LL}, left).

The first term in Eq.~\eqref{eq:mainPrMatch} is to ensure that the classifier continues to perform well on the internal distribution (note that the pseudo-dataset  approximates this distribution). The second term is the domain alignment  matching loss which enforces the target domain to share the internal distribution in the embedding space to make the embedding domain-invariant.  The major remaining question is selection of the distribution metric.
We choose  SWD metric~\cite{courty2017optimal} to computer $D(\cdot,\cdot)$ due to its suitability for deep learning due to possessing non-vanishing gradients when two distributions have non-overlapping supports~\cite{bonnotte2013unidimensional,lee2019sliced}.
SWD inherits this property from WD, yet the advantage of using SWD over WD is that SWD can be computed efficiently using a closed form solution. Additionally, empirical version of SWD can be computed using the samples that are drawn from the corresponding two distributions, as it is the case in Eq.~\eqref{eq:mainPrMatch}.
 Hence, Eq.~\eqref{eq:mainPrMatch} can be solved using first-order optimization techniques (see the Appendix   for more details on properties of SWD). Our proposed solution, named Sequential Model Adaptation Using Internal distribution (SMAUI), is presented  and visualized in Algorithm~\ref{NeuripsUDAalgorithm}.  Figure~\ref{figDL:LL} 
 presents the high-level description of  SMAUI.

 \begin{algorithm}[H]
 \small
\caption{$\mathrm{SDAUP}\left (\lambda , ITR \right)$\label{NeuripsUDAalgorithm}} 
 {\small
\begin{algorithmic}[1]
\State \textbf{Initial Training}: 
\State \hspace{2mm}\textbf{Input:} source dataset $\mathcal{D}_{\mathcal{S}}=(\bm{X}_{\mathcal{S}},  \bm{Y}_{\mathcal{S}})$,
\State \hspace{4mm}\textbf{Training on Source Domain:}
\State \hspace{4mm} $\hat{ \theta}_0=(\hat{\bm{w}}_0,\hat{\bm{v}}_0) =\arg\min_{\theta}\sum_i \mathcal{L}(f_{\theta}(\bm{x}_i^s),\bm{y}_i^s)$
\State \hspace{2mm}  \textbf{Internal Distribution Estimation:}
\State \hspace{4mm} Use Eq.~\eqref{eq:MAPest} and estimate $\alpha_j, \bm{\mu}_j,$ and $\Sigma_j$
\State \textbf{Model Adaptation}: 
\State \hspace{2mm} \textbf{Input:} target dataset $\mathcal{D}_{\mathcal{T}}=(\bm{X}_{\mathcal{S}})$
\State \hspace{2mm} \textbf{Pseudo-Dataset Generation:} 
\State \hspace{4mm} $\mathcal{\hat{D}}_{\mathcal{P}}=(\textbf{Z}_{\mathcal{P}},\textbf{Y}_{\mathcal{P}})=$
\State \hspace{12mm} $([\bm{z}_1^p,\ldots,\bm{z}_N^p],[\bm{y}_1^p,\ldots,\bm{y}_N^p])$, where:
\State \hspace{16mm} $\bm{z}_i^p\sim \hat{p}_J(\bm{z}), 1\le i\le N_p$
\State \hspace{17mm}$\bm{y}_i^p=
\arg\max_j\{h_{\hat{\bm{w}}_0}(\bm{z}_i^p)\}$
\For{$itr = 1,\ldots, ITR$ }
\State draw data batches from $\mathcal{\hat{D}}_{\mathcal{T}}$ and $\mathcal{\hat{D}}_{\mathcal{P}}$
\State Update the model by solving Eq.~\eqref{eq:mainPrMatch}
\EndFor
\end{algorithmic}}
\end{algorithm}

\section{Theoretical Analysis}
We demonstrate that our algorithm optimizes an upperbound for the target domain expected risk. We treat the embedding space as the hypothesis space within the standard PAC-learning in our analysis. We consider the hypothesis class for PAC-learning to be the set of classifier subnetworks parameterized by $\bm{w}$, $\mathcal{H} = \{h_{\bm{w}}(\cdot)|h_{\bm{w}}(\cdot):\mathcal{Z}\rightarrow \mathbb{R}^k, \bm{v}\in \mathbb{R}^V\}$. We denote the observed risk on the source and the target domains by $e_{\mathcal{S}}$ and  $e_{\mathcal{T}}$, respectively. Also, let $\hat{\mu}_{\mathcal{S}}=\frac{1}{N}\sum_{n=1}^N\delta(\phi_{\bm{v}}(\bm{x}_n^s))$ and $\hat{\mu}_{\mathcal{T}}=\frac{1}{M}\sum_{m=1}^M\delta(\phi_{\bm{v}}(\bm{x}_m^t))$ denote the empirical source and the empirical target distributions in the embedding space. These distributions are computed using the representations of the training data in the embedding space. Similarly, let $\hat{\mu}_{\mathcal{P}}=\frac{1}{N_p}\sum_{q=1}^{N_p}\delta(\bm{z}_n^q)$ denote the empirical internal distribution built using the pseudo-dataset.  Moreover, let $h_{\bm{w}^*}$ denote the optimal model  that minimizes the combined source and target risks $e_{\mathcal{C}}(\bm{w}^*)$, i.e. $\bm{w}^*= \arg\min_{\bm{w}} e_{\mathcal{C}}(\bm{w})=\arg\min_{\bm{w}}\{ e_{\mathcal{S}}+  e_{\mathcal{T}}\}$. In the presence of enough labeled target domain data, this is  the best  model that can be learned jointly on both domains.   Finally, since we generate a clean pseudo-dataset, we can let $\tau =  \mathbb{E}_{\bm{z}\sim \hat{p_{J}(\bm{z})}}(\mathcal{L}(h(\bm{z}),h_{\hat{\bm{w}}_0}(\bm{z})) $ as the expected risk  of the optimal model for the source domain data on the generated pseudo-dataset using the GMM. We conclude: 


\textbf{Theorem 1}: Consider that we generate a pseudo-dataset  using the estimate for the internally learned distribution and update the model sequentially using the algorithm~\ref{NeuripsUDAalgorithm}. Then, the following holds:
\begin{equation}
\small
\begin{split}
&e_{\mathcal{T}}\le  e_{\mathcal{S}} +W(\hat{\mu}_{\mathcal{S}},\hat{\mu}_{\mathcal{P}})+W(\hat{\mu}_{\mathcal{T}},\hat{\mu}_{\mathcal{P}})+(1-\tau)+e_{\mathcal{C'}}(\bm{w}^*)\\&+\sqrt{\big(2\log(\frac{1}{\xi})/\zeta\big)}\big(\sqrt{\frac{1}{N}}+\sqrt{\frac{1}{M }}+2\sqrt{\frac{1}{N_p }}\big),
\end{split}
\label{eq:theroemforPLnips}
\end{equation}    
where $W(\cdot,\cdot)$ denotes the WD distance and   $\xi$ is a constant.

\textbf{Proof:}  The complete proof  is included in  the Appendix.

  We can use Theorem~1 to  explain  why our algorithm can adapt the model that is trained using the source domain to generalize well on the target domain. We can see that SMAUI algorithm  minimizes the upperbound of the target domain risk as given in Eq.~\eqref{eq:theroemforPLnips}. The first three terms are minimized directly in our optimization. The source risk $e_{\mathcal{S}}$ is minimized through the initial training on the source domain prior to the model adaptation step. We minimize the second term in the upperbound of Eq.~\eqref{eq:theroemforPLnips}  by intentionally fitting a GMM distribution on the source domain distribution in the embedding. We note that this term is small  if the source domain distribution can be fit well with a GMM distribution in reality. Hence, this is a constraint for applicability of our method. Note, however, all the parametric learning methods face a similar limitation and this is not particular to our algorithm.  The third term is minimized directly because it is  the second term in the objective function of Eq.~\eqref{eq:mainPrMatch}. 
  The fourth and fifth terms are not minimized by our algorithm, rather state conditions under which the algorithm would work. 
  The term $(1-\tau)$  is small if we set $\tau\approx 1$ to generate a clean pseudo-dataset and cancel out the outliers. We note that if $\tau$ is chosen very close to 1, then the pseudo-dataset samples would all be close to the mean and hence pseudo-dataset will not capture the higher-order moments of the internal distribution.  
  The term $e_{C'}(\bm{w}^*)$ is   a constant term. This term will be small if the domains are related, i.e., share the same classes with the same consistent label-space encoding, and in presence of the target   labeled data, the base model can be trained to work well on both domains. This means that aligning the distributions in the embedding must be a possibility for our algorithm to work. Finally, the last term in Eq.~\eqref{eq:theroemforPLnips} is a constant term that merely states that in order to train a good model, we need sufficiently large source and target datasets and also we need to generate a large pseudo-dataset.   We conclude   assuming that the   domains are related and applicability of the GMM estimation for the source distribution, SMAUI minimizes all the remaining terms in the upperbound of Eq.~\eqref{eq:theroemforPLnips}.
  

 \section{Experimental Validation}
We compare our algorithm against several recently developed UDA algorithms using benchmark   UDA tasks due to closeness of the UDA learning setting to the sequential model adaptation setting. 
Our   code  is provided as a supplement. 

 \textbf{Datasets:}
We validate our method on five  standard UDA benchmarks and adapted them for sequential task learning: \textbf{Digit recognition   tasks}, \textbf{Office-31 Detest}, \textbf{ImageCLEF-DA Dataset}, \textbf{Office-Caltech Dataset}, and \textbf{VisDA-2017}. Details about these datasets are included in the Appendix. 

\textbf{Empirical Evaluation:} 
Since sequential model adaptation is not a well-explored problem, we follow the UDA literature for evaluation due to the topic proximity. For this reason, we use the metrics and the features used in the UDA literature for fair comparison.  We use the VGG16 network  as the base model for the digit recognition tasks. The network is initialized with random weights. We use  the Decaf6 features  for the Office-Caltech tasks. For the Office-31 and ImageCLEF-DA datasets, we use  the  ResNet-50 backbone which is pre-trained on the ImageNet. For the VisDA-2017 dataset, we use the  ResNet-101 backbone pre-trained on the ImageNet.
 A point of strength for our algorithm is that there are only two major algorithm-specific hyper-parameters and tuning them is not challenging. We set $\tau=0.99$ and $\lambda = 10^{-3}$. 


We report the average classification rate on the target domain and the standard deviation based on ten runs for each UDA task. We train the base model using the source labeled data. We report the performance of the model before adaptation  as a baseline which also demonstrates the effect of domain shift. Then we adapt the model using the target unlabeled data using SMAUI algorithm and report the     target domain performance.  In our Tables, bold font denotes the best performance. The baseline performance before model adaptation is reported in the first row, then the UDA methods based on adversarial learning, then the UDA methods based on direct matching which are separated by a line in the middle, followed by our result in the last rows of the tables.  

 
 We compare our method against methods that are based on adversarial learning: GtA~\cite{sankaranarayanan2018generate}, DANN~\cite{ganin2016domain}, ADDA~\cite{tzeng2017adversarial}, MADA~\cite{pei2018multi},  SymNets~\cite{zhang2019domain},   CDAN~\cite{long2018conditional},   MMAN~\cite{ma2019deep}, and DANCE~\cite{saito2020universal}. We have also included  methods based on direct distribution matching:  DAN~\cite{long2015learning}, DRCN~\cite{ghifary2016deep},
CORAL~\cite{sun2016return}, RevGrad~\cite{ganin2014unsupervised}, CAN~\cite{kang2019contrastive}, JAN~\cite{long2017deep},  WDGRL~\cite{shen2018wasserstein},   JDDA~\cite{chen2019joint},   ETD~\cite{li2020enhanced}, and SRADA~\cite{wang2020self}.   
Finally, we also compared against UDAwSD~\cite{li2020model} and SHOT~\cite{liang2020we}  which are  source-free UDA methods. 
For each dataset, we include results of the works for which the original paper has used that dataset and reported the corresponding performance. 
For more details on the experimental setup, please refer to the Appendix.

  \begin{table*}[t!]
 \setlength{\tabcolsep}{3pt}
 \centering 
{\small
\begin{tabular}{lc|ccc|c|ccc}   
\multicolumn{2}{c}{Method}    & $\mathcal{M}\rightarrow\mathcal{U}$ & $\mathcal{U}\rightarrow\mathcal{M}$ & $\mathcal{S}\rightarrow\mathcal{M}$ &Method & $\mathcal{M}\rightarrow\mathcal{U}$ & $\mathcal{U}\rightarrow\mathcal{M}$ & $\mathcal{S}\rightarrow\mathcal{M}$\\
\hline
\multicolumn{2}{c|}{GtA }& 92.8  $\pm$  0.9	&	90.8  $\pm$  1.3	&	92.4  $\pm$  0.9 &CDAN &93.9 &96.9& 88.5 \\
\multicolumn{2}{c|}{ADDA }& 89.4  $\pm$  0.2&90.1  $\pm$  0.8&76.0  $\pm$  1.8& CyCADA &   95.6   $\pm$  0.2  & 96.5  $\pm$  0.1 & 90.4  $\pm$  0.4  \\  
\multicolumn{2}{c|}{SRADA }& 94.1  &98.0&-& -&   -& - & - \\  
\hline
\multicolumn{2}{c|}{RevGrad~ }&	 77.1  $\pm$  1.8 	&	73.0  $\pm$  2.0 	&	73.9  &JDDA & -& 97.0 $\pm$0.2 &  93.1  $\pm$0.2   \\ 
\multicolumn{2}{c|}{DRCN }&	 91.8  $\pm$  0.1 	&	73.7  $\pm$  0.4 	&	82.0  $\pm$  0.2 &  OPDA & 70.0 & 60.2 &-   \\
\multicolumn{2}{c|}{ETD }& \textbf{96.4}$\pm$ 0.3 & 96.3$\pm$ 0.1&\textbf{97.9}$\pm$ 0.4&
MML & 77.9  & 60.5  &62.9  \\ 
\hline
\multicolumn{2}{c|}{Source Only}&	 90.1$\pm$2.6	&	80.2$\pm$5.7	&	67.3$\pm$2.6   & SMAUI &   92.2  $\pm$  0.5	&	\textbf{98.2}  $\pm$  0.2	&	92.6  $\pm$  1.0 \\
\end{tabular}}
\caption{ Classification accuracy for   MINIST, USPS, and SVHN  digit recognition   datasets.    }
\label{table:tabDA1}
\vspace{-2mm}
 \end{table*}

  \begin{table*}[t!]
 \centering 
{\footnotesize
\begin{tabular}{lc|cccccc|c}   
\multicolumn{2}{c}{Method}    & $\mathcal{A}\rightarrow\mathcal{W}$ & $\mathcal{D}\rightarrow\mathcal{W}$ & $\mathcal{W}\rightarrow\mathcal{D}$ &$\mathcal{A}\rightarrow\mathcal{D}$ &$\mathcal{D}\rightarrow\mathcal{A}$ &$\mathcal{W}\rightarrow\mathcal{A}$ & Average\\
\hline
\multicolumn{2}{c|}{Source Only }&68.4  $\pm$  0.2 & 96.7  $\pm$  0.1&  99.3  $\pm$  0.1&  68.9  $\pm$  0.2 & 62.5  $\pm$  0.3&  60.7  $\pm$  0.3  & 75.6\\
\hline
\multicolumn{2}{c|}{GtA }&89.5  $\pm$  0.5& 97.9  $\pm$  0.3& 99.8  $\pm$  0.4& 87.7  $\pm$  0.5& 72.8  $\pm$  0.3& 71.4  $\pm$  0.4&86.5 \\
\multicolumn{2}{c|}{DANN }&   82.0  $\pm$  0.4& 96.9  $\pm$  0.2& 99.1  $\pm$  0.1& 79.7  $\pm$  0.4& 68.2  $\pm$  0.4 &67.4  $\pm$  0.5 & 82.2 \\ 
\multicolumn{2}{c|}{ADDA }& 86.2  $\pm$  0.5 & 96.2  $\pm$  0.3&  98.4  $\pm$  0.3&  77.8  $\pm$  0.3 & 69.5  $\pm$  0.4&  68.9  $\pm$  0.5& 82.8 \\
\multicolumn{2}{c|}{SymNets }& 90.8  $\pm$  0.1& 98.8 $\pm$  0.3& \textbf{100.0}  $\pm$  .0& 93.9  $\pm$  0.5&  74.6  $\pm$  0.6& 72.5  $\pm$  0.5&88.4 \\  
\multicolumn{2}{c|}{MADA}& 82.0  $\pm$  0.4&  96.9  $\pm$  0.2 & 99.1  $\pm$  0.1&  79.7  $\pm$  0.4 & 68.2  $\pm$  0.4&  67.4  $\pm$  0.5 & 82.2  \\
\multicolumn{2}{c|}{CDAN }&93.1  $\pm$  0.2 &98.2  $\pm$  0.2& \textbf{100.0}  $\pm$  0.0& 89.8  $\pm$  0.3& 70.1 $\pm$  0.4& 68.0  $\pm$  0.4& 86.6 \\  
\multicolumn{2}{c|}{SRADA }&95.2 &98.6 &\textbf{100.0} &91.7 &74.5 &73.7 &89.0 \\  
\multicolumn{2}{c|}{UDAwSD }&	 93.7$\pm$0.2 &98.5$\pm$0.1 & 99.8 $\pm$0.2&92.7$\pm$0.4& 75.3 $\pm$0.5&${\textbf{77.8}}$$\pm$0.1 &89.6 \\  
\hline
\multicolumn{2}{c|}{DAN }&  80.5 $\pm$ 0.4 &97.1 $\pm$ 0.2& 99.6 $\pm$ 0.1& 78.6 $\pm$ 0.2& 63.6 $\pm$ 0.3& 62.8 $\pm$ 0.2& 80.4\\ 
\multicolumn{2}{c|}{DRCN }&	 72.6  $\pm$  0.3 	&	96.4  $\pm$  0.1	& 99.2  $\pm$  0.3 	& 67.1  $\pm$  0.3 	& 56.0  $\pm$  0.5 	&72.6 $\pm$  0.3 	& 77.7  \\ 
\multicolumn{2}{c|}{RevGrad } &82.0  $\pm$  0.4&96.9  $\pm$  0.2& 99.1  $\pm$  0.1& 79.7  $\pm$  0.4& 68.2  $\pm$  0.4& 67.4  $\pm$  0.5 & 82.2 \\
\multicolumn{2}{c|}{CAN }&94.5 $\pm$  0.3& 99.1 $\pm$  0.2& 99.8  $\pm$  0.2& 95.0 $\pm$  0.3& \textbf{78.0}  $\pm$  0.3& 77.0 $\pm$  0.3 & \textbf{90.6}\\ 
\multicolumn{2}{c|}{JAN }& 85.4  $\pm$  0.3& 97.4  $\pm$  0.2& 99.8  $\pm$  0.2& 84.7  $\pm$  0.3& 68.6  $\pm$  0.3 &70.0  $\pm$  0.4  & 85.8\\ 
\multicolumn{2}{c|}{JDDA }&82.6  $\pm$  0.4& 95.2  $\pm$  0.2& 99.7  $\pm$  0.0 &79.8  $\pm$  0.1 &57.4  $\pm$  0.0& 66.7  $\pm$  0.2 & 80.2\\ %
\multicolumn{2}{c|}{ETD }&92.1&\textbf{100.0}&  \textbf{100.0}&88.0&71.0&67.8&  86.2\\
\multicolumn{2}{c|}{DANCE }&88.6& 97.5& \textbf{100.0}& 89.4 &69.5 &68.2 &85.5 \\
\multicolumn{2}{c|}{SHOT }&91.2&98.3&99.9&90.6&72.5&71.4&87.3\\
  \hline
\multicolumn{2}{c|}{SMAUI}&	 \textbf{97.8}  $\pm$  2.1	&	95.6  $\pm$  0.5	&	99.1  $\pm$  0.3 & \textbf{97.8}  $\pm$  1.7   & 68.2  $\pm$  4.5   & 71.7  $\pm$  3.6 & 88.4\\   
\end{tabular}}
\caption{ Classification accuracy for UDA tasks for  Office-31 dataset. }
\label{table:tabDA2}
\vspace{-2mm}
 \end{table*}

  \begin{table*}[t!]
 \centering 
{\footnotesize
\begin{tabular}{lc|cccccc|c}   
\multicolumn{2}{c}{Method}    & $\mathcal{I}\rightarrow\mathcal{P}$ & $\mathcal{P}\rightarrow\mathcal{I}$ & $\mathcal{I}\rightarrow\mathcal{C}$ &$\mathcal{C}\rightarrow\mathcal{I}$ &$\mathcal{C}\rightarrow\mathcal{P}$ &$\mathcal{P}\rightarrow\mathcal{C}$& Average \\
\hline
\multicolumn{2}{c|}{Source Only }& 74.8  $\pm$  0.3& 83.9  $\pm$  0.1& 91.5  $\pm$  0.3 &78.0  $\pm$  0.2 &65.5  $\pm$  0.3& 91.2  $\pm$  0.3&80.8  \\
\hline
\multicolumn{2}{c|}{DANN }&   82.0  $\pm$  0.4& 96.9  $\pm$  0.2& 99.1  $\pm$  0.1& 79.7  $\pm$  0.4& 68.2  $\pm$  0.4 &67.4  $\pm$  0.5 &82.2 \\ 
\multicolumn{2}{c|}{SymNets }& 80.2 $\pm$ 0.3 &93.6 $\pm$ 0.2& 97.0 $\pm$ 0.3 &93.4 $\pm$ 0.3 &78.7 $\pm$ 0.3 &96.4 $\pm$ 0.1 &89.9\\  
\multicolumn{2}{c|}{MADA }& 75.0 $\pm$ 0.3& 87.9 $\pm$ 0.2& 96.0 $\pm$ 0.3& 88.8 $\pm$ 0.3& 75.2 $\pm$ 0.2& 92.2 $\pm$ 0.3   & 85.9 \\
\multicolumn{2}{c|}{CDAN  }&76.7 $\pm$ 0.3& 90.6 $\pm$ 0.3& 97.0 $\pm$ 0.4 &90.5 $\pm$ 0.4& 74.5 $\pm$ 0.3 &93.5 $\pm$ 0.4 &87.1\\  
\multicolumn{2}{c|}{SRADA }&78.3& 91.3 &96.7 &90.5 &78.1& 96.2& 88.5\\  
\hline
\multicolumn{2}{c|}{DAN }& 74.5 $\pm$ 0.4 &82.2 $\pm$ 0.2 &92.8 $\pm$ 0.2& 86.3 $\pm$ 0.4& 69.2 $\pm$ 0.4& 89.8 $\pm$ 0.4&82.4\\ 
\multicolumn{2}{c|}{RevGrad } &75.0 $\pm$ 0.6 &86.0 $\pm$ 0.3& 96.2 $\pm$ 0.4 &87.0 $\pm$ 0.5& 74.3 $\pm$ 0.5& 91.5 $\pm$ 0.6&85.0 \\
\multicolumn{2}{c|}{JAN }&  76.8 $\pm$ 0.4&  88.0 $\pm$ 0.2&  94.7 $\pm$ 0.2 & 89.5 $\pm$ 0.3&  74.2 $\pm$ 0.3 & 91.7 $\pm$ 0.3&85.7\\ %
\multicolumn{2}{c|}{ETD }&  81.0 & 91.7 & 97.9 & 93.3 & 79.5 & 95.0&  89.7 \\ 
\hline
\multicolumn{2}{c|}{SMAUI}&	 		\textbf{88.7}  $\pm$  1.2	&	\textbf{99.5}  $\pm$  0.2 & \textbf{100}  $\pm$  0.0   & \textbf{94.9}  $\pm$  0.3 & \textbf{88.8}  $\pm$  0.9 & \textbf{99.8}  $\pm$  0.0     & \textbf{95.3} \\
\end{tabular}}
\caption{ Classification accuracy for UDA tasks for  ImageCLEF-DA dataset. }
\label{table:tabDA3}
\vspace{-2mm}
 \end{table*}

  \begin{table*}[t!]
 \centering 
 \setlength{\tabcolsep}{2pt}
{\footnotesize
\begin{tabular}{lc|cccccccccccc|c}   
\multicolumn{2}{c}{Method}    &A$\rightarrow$C & A$\rightarrow$D & A$\rightarrow$W&W$\rightarrow$A&W$\rightarrow$D &W$\rightarrow$C &D$\rightarrow$A & D$\rightarrow$W & D$\rightarrow$C&C$\rightarrow$A&C$\rightarrow$W &C$\rightarrow$D& Average\\
\hline
\multicolumn{2}{c|}{Source Only}& 84.6 &81.1 &75.6 &79.8 &98.3 &79.6 &84.6 &96.8& 80.5 &92.4 &84.2& 87.7& 85.4 \\
\hline
\multicolumn{2}{c|}{DANN }&  87.8&  82.5&  77.8&  83.0&  100 & 81.3&  84.7&  99.0 & 82.1 & 93.3 & 89.5&  91.2&  87.7   \\
\multicolumn{2}{c|}{MMAN }& 88.7 &97.5& 96.6& 94.2& 100& 89.4& \textbf{94.3}& 99.3& 87.9& 93.7& 98.3 &\textbf{98.1}& 94.6 \\
\hline
\multicolumn{2}{c|}{RevGrad }&85.7 &89.2& 90.8 &93.8& 98.7& 86.9& 90.6 &98.3 &83.7 &92.8& 88.1 &87.9& 88.9 \\
\multicolumn{2}{c|}{DAN }&84.1 &91.7& 91.8 &92.1 &100& 81.2 &90.0 &\textbf{98.5} &80.3 &92.0 &90.6& 89.3& 90.1\\
\multicolumn{2}{c|}{CORAL }&  86.2 &91.2 &90.5 &88.4 &100 &88.6 &85.8 &97.9& 85.4 &93.0 &92.6 &89.5& 90.8\\  
\multicolumn{2}{c|}{WDGR }& 87.0& 93.7 &89.5 &93.7 &100& 89.4 &91.7 &97.9& 90.2 &93.5& 91.6 &94.7 &92.7 \\  
\hline
\multicolumn{2}{c|}{SMAUI}&  \textbf{99.9} &\textbf{100.0} & \textbf{96.7} & \textbf{96.8}&  \textbf{100}& \textbf{94.4}& 84.8 & 93.4 &  \textbf{91.8} & \textbf{98.8} & 80.4  & 91.4 &  \textbf{94.0}
\end{tabular}}
\caption{ Performance comparison  for UDA tasks of  Office-Caltech dataset. }
\label{table:tabDA21}
\vspace{-2mm}
 \end{table*}

 \begin{table}[t!]
 \centering 
 \setlength{\tabcolsep}{2pt}
{\scriptsize
\begin{tabular}{lc|cccccccc}   
\multicolumn{2}{c}{Task}    & JAN& DJT & GtA & SimNet & CDAN &DANCE 
 & MCD  & SMAUI \\
\hline
\multicolumn{2}{c|}{Syn.$\rightarrow$Real.}& 61.6 & 66.9	&69.5	&69.6  & 70.0 &70.2& 71.9 & \textbf{76.9}  $\pm$  0.7  \\
\end{tabular}}
\caption{ Classification accuracy for the VisDA UDA task. }
\label{table:tabDA5}\vspace{-4mm}
 \end{table}


 \begin{figure*}[tbh]
    \centering
           \begin{subfigure}[b]{0.23\textwidth}\includegraphics[width=\textwidth]{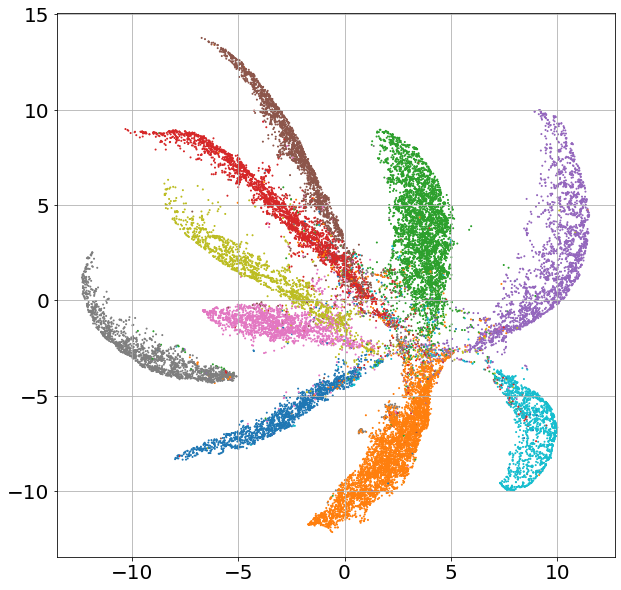}
           \centering
        \caption{ }
        \label{NIPSDALfig:MNISTUSPS}
    \end{subfigure}
    \begin{subfigure}[b]{0.23\textwidth}\includegraphics[width=\textwidth]{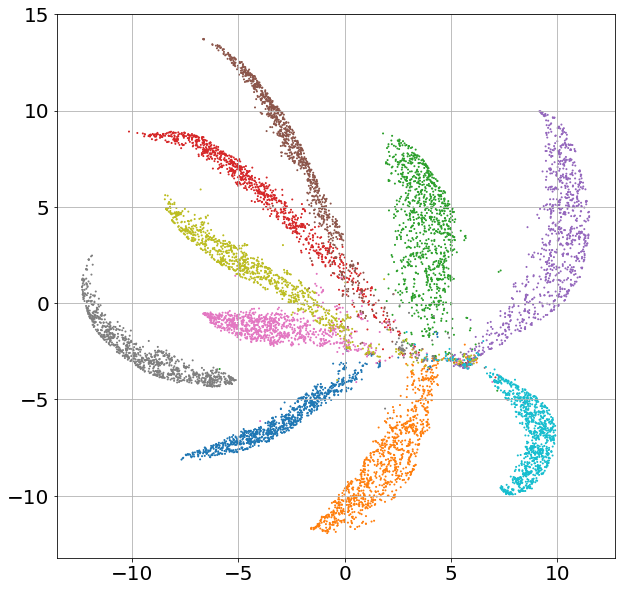}
           \centering
        \caption{ }
        \label{NIPSDALfig:USPSMNIST}
    \end{subfigure}
       \begin{subfigure}[b]{0.23\textwidth}\includegraphics[width=\textwidth]{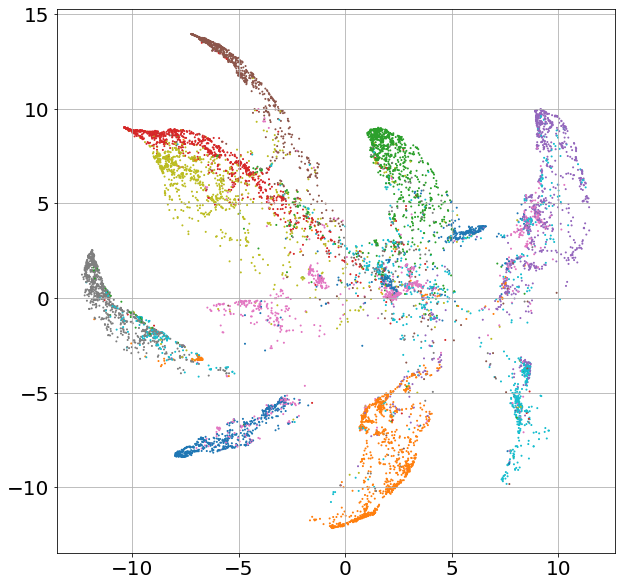}
           \centering
        \caption{ }
        \label{NIPSDALfig:MNISTUSPSembed}
    \end{subfigure}
       \begin{subfigure}[b]{0.23\textwidth}\includegraphics[width=\textwidth]{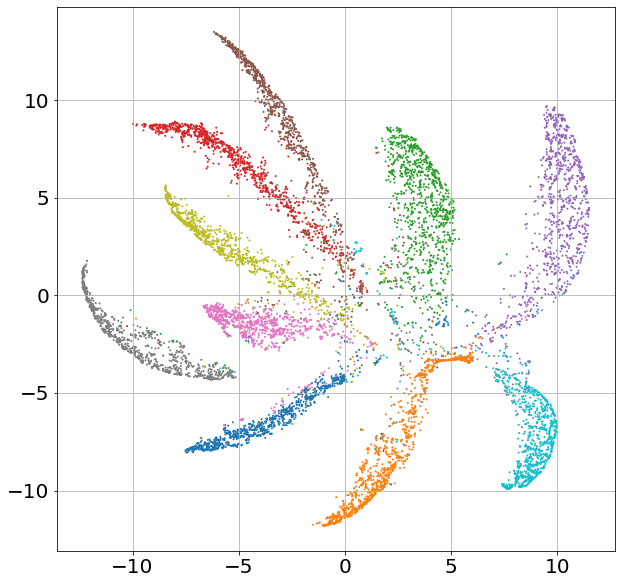}
              \centering
        \caption{ }
        \label{NIPSDALfig:USPSMNISTembed}
    \end{subfigure}
     \caption{\small UMAP visualization for  the $\mathcal{S}\rightarrow \mathcal{M}$ task:  (a) the source domain   testing split, (b) the internal distribution samples, (c) the target domain testing split prior to adaptation, and (d) post adaptation.  (Best viewed in color).  }\label{NIPSDALfig:resultsCatforgetRelated}
\end{figure*}

\begin{figure*}[tb!]
  \centering
    \begin{subfigure}[b]{0.19\textwidth}\includegraphics[width=\textwidth]{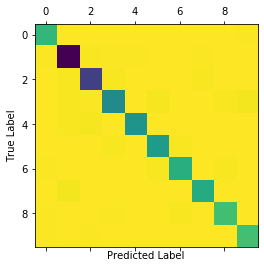}
        \caption{ }
        \label{NIPSDALfig:CMsource}
    \end{subfigure}
  \centering
    \begin{subfigure}[b]{0.19\textwidth}\includegraphics[width=\textwidth]{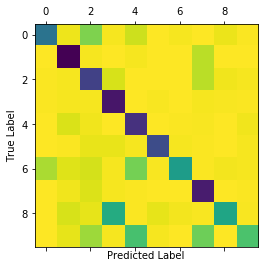}
        \caption{ }
        \label{NIPSDALfig:CMbefore}
    \end{subfigure}
       \begin{subfigure}[b]{0.19\textwidth}\includegraphics[width=\textwidth]{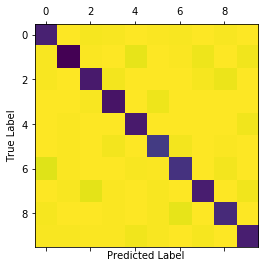}
           \centering
       \caption{ }
        \label{NIPSDALfig:CMafter}
    \end{subfigure}
      \centering
           \begin{subfigure}[b]{0.19\textwidth}\includegraphics[width=\textwidth]{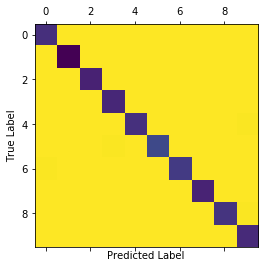}
           \centering
        \caption{ }
        \label{NIPSDALfig:CMtarget}
    \end{subfigure}
       \begin{subfigure}[b]{0.19\textwidth}\includegraphics[width=\textwidth]{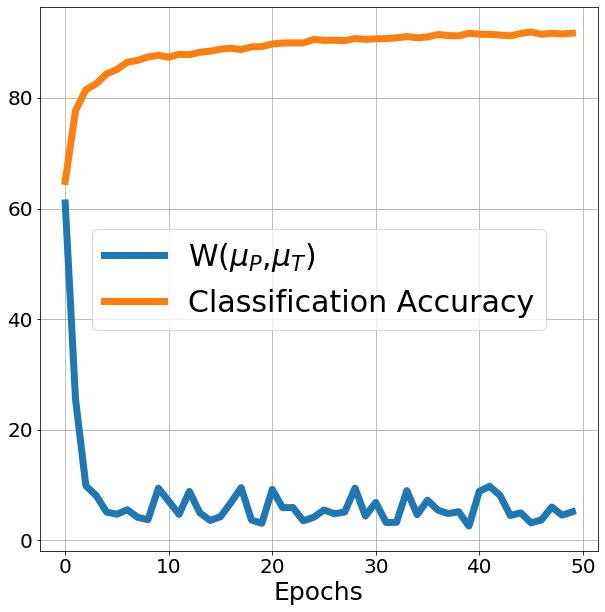}
              \centering
       \caption{ }
        \label{NIPSDALfig:converg}
    \end{subfigure}
     \caption{\small Confusion matrices for  the $\mathcal{S}\rightarrow \mathcal{M}$ task:  (a) the source domain (b) the target domain prior to adaptation, (c) the target domain after adaptation, (d) the target domain with a model trained using the target   labeled data,  (e) the cross-domain distribution \& the test error vs epochs.    }\label{NIPSDALfig:analysis}
\end{figure*}

\subsection{Results}

Results for the digit recognition tasks are reported in Table~\ref{table:tabDA1}. Despite the sequential training regime, we observe SMAUI outperforms the other methods in one of these tasks and its performance is quite competitive in the other two tasks.  SMAUI leads to strong results compared to the joint UDA methods because stabilizing the internal distribution for both domain would   mitigate   domain shift.
 We also observe that performance of the methods based on direct probability matching has improved recently which might because of using secondary mechanisms to improve generalization.
 
Table~\ref{table:tabDA2} summarizes the results for  Office-31 dataset. We see in two of the tasks SMAUI leads to the best results and for the rest of the tasks is still competitive. Note, however, it seems there is no clear winner algorithm across all the tasks of this dataset. This maybe because it is known that some labels in this dataset are noisy and some images contain objects that belong to other classes~\cite{bousmalis2016domain}. The approach we use to align the two distributions  is vulnerable with respect to this label pollution.
We conclude that label pollution can degrade the performance of SMAUI.   

Results for UDA tasks of the ImageCLEF-DA dataset are reported in Table~\ref{table:tabDA3}. We see that although SMAUI does not use the source samples during model adaptation, it leads to a significant performance boost over the prior methods for this dataset. This may be because
the Caltech-256, the   ILSVRC 2012, and the Pascal VOC 2012  datasets have the equal size and the domains are balanced across the classes. As a result, matching the source and the target distributions to the same internal multi-modal distribution is more straightforward for the ImageCLEF-DA  dataset. 
Since we rely on empirical distributions for alignment, balanced datasets represent true distributions better.
We have reported results for the Office-Caltech dataset in Table~\ref{table:tabDA21}.
We note that SMAUI leads to a relatively   competitive performance on the tasks and in terms of average performance outperforms the listed UDA methods in this dataset. We conclude that  addition of the Caltech-256 dataset might have helped mitigating the effect of label pollution which  has improved   our performance.

Results for VisDA task is presented in Table~\ref{table:tabDA5}.
We observe a significant boost in performance for VisDA task.
From inspecting Tables~1--5, we conclude that there is no single UDA  method with the best performance on all the tasks. This is natural because these datasets are diverse in terms of difficulty, cross-domain gap, dataset size, label-pollution, etc, and any particular method may be more vulnerable in special cases. However, we note that although these UDA methods should serve an upperbound for SMAUI, SMAUI  works reasonably well on all the domain adaptation tasks. 
These results demonstrate that although our motivation was to address sequential model adaptation, SMAUI can also be used as a standard joint training UDA algorithm with competitive results. Moreover, it can also be used as a source-free domain adaptation algorithm for preserving privacy~\cite{stan2021privacy} and multi-source UDA \cite{stan2021secure}.



\subsection{Analysis and Ablative Studies}
To demonstrate that SMAUI algorithm implements what we anticipated and to provide a better intuition about its effectiveness, we  have used the UMAP~\cite{mcinnes2018umap} visualization tool to reduce  the dimension of the data representations  in the embedding space to two for 2D visualization purpose. Figure~\ref{NIPSDALfig:resultsCatforgetRelated} represents the testing data splits of the source and the target domains and samples of the internal distribution for the $\mathcal{S}\rightarrow \mathcal{M}$ digit recognition task. In this figure, each point represents one data point and each color represents one of the ten digit classes. Comparing Figures~\ref{NIPSDALfig:MNISTUSPS}  and \ref{NIPSDALfig:USPSMNIST}, we can see that the high-confidence internal GMM   distribution samples approximate the source domain distribution reasonably well. Figure~\ref{NIPSDALfig:MNISTUSPSembed} denotes that the target domain samples are separable prior to adaptation to some extent due to domain similarity, but we can observe more regions with overlapped classes, i.e., less separability, that lead to performance degradation. This is due to distributional gap between the two domains. Figure~\ref{NIPSDALfig:USPSMNISTembed} denotes that SMAUI algorithm has successfully  aligned the source and the target distributions using the intermediate internal distribution. Figure~\ref{NIPSDALfig:resultsCatforgetRelated} empirically confirms the results that we  deduced form  Theorem~1 because we observe that the upperbound minimization has led to improved generalization on the target domain.

For a class-level analysis of effect of model adaptation, Figures~\ref{NIPSDALfig:CMsource}--3d visualize the confusion matrices for the classifier with explanations in the caption. We can see in Figure~\ref{NIPSDALfig:CMbefore} that domain shift causes confusion   between digit classes that are in visually similar classes, e.g., digits ``3'' and ``8'' or digits ``4'', ``7'', and ``9''. As seen in  Figure~\ref{NIPSDALfig:CMafter},  the confusion is reduced for all classes using SMAUI algorithm. Comparing Figure~\ref{NIPSDALfig:CMafter} with Figures~\ref{NIPSDALfig:CMsource} and 3d, we see that the initial confusions between the classes in the source domain translate into the   target domain, despite the  fact that the target domain ($\mathcal{M}$) is an easier problem (Figures~\ref{NIPSDALfig:CMtarget}). This observation is predictable from Theorem~1 because the model performance on the target domain is upperbounded by the source domain error, even if it is a simpler problem. We again conclude that empirical results support our theorem.

We have also validated Theorem~1 in terms of effect of SMAUI on performance in the target domain in Figure~\ref{NIPSDALfig:converg}. We have plotted the test error versus optimization epochs during model adaptation process. We observe that as more training epochs are performed and the distributions are aligned progressively, i.e., domain discrepancy is minimized, the testing accuracy on the target domain accuracy constantly increases. This observation accords with theoretical prediction by Eq.~\eqref{eq:theroemforPLnips} because as SMAUI gradually aligns the target domain distribution with the source   distribution using the internal distribution,  the upperbound in Eq.~\eqref{eq:theroemforPLnips} becomes tighter.


 \section{Conclusions  }
 We  addressed the problem of model adaptation in a sequential task learning setting. Our algorithm is based on minimizing the distributional domain discrepancy in a shared embedding space using an intermediate multi-modal internal distributional GMM estimate. We estimate the source domain internal distribution using a GMM distribution. The internal distribution encodes what has been learned from the source domain.  As a result, we can adapt the source-trained classifier using this distribution as an intermediate cross-domain distribution     to generalize better on the target domain.      A  future   direction is equipping the algorithm   with class-conditional alignment mechanisms to improve its performance.

\bibliography{AAAI}

\begin{thebibliography}{57}
\providecommand{\natexlab}[1]{#1}

\bibitem[{Bhushan~Damodaran et~al.(2018)Bhushan~Damodaran, Kellenberger,
  Flamary, Tuia, and Courty}]{damodaran2018deepjdot}
Bhushan~Damodaran, B.; Kellenberger, B.; Flamary, R.; Tuia, D.; and Courty, N.
  2018.
\newblock Deepjdot: Deep joint distribution optimal transport for unsupervised
  domain adaptation.
\newblock In \emph{Proceedings of the European Conference on Computer Vision
  (ECCV)}, 447--463.

\bibitem[{Bolley, Guillin, and Villani(2007)}]{bolley2007quantitative}
Bolley, F.; Guillin, A.; and Villani, C. 2007.
\newblock Quantitative concentration inequalities for empirical measures on
  non-compact spaces.
\newblock \emph{Probability Theory and Related Fields}, 137(3-4): 541--593.

\bibitem[{Bonnotte(2013)}]{bonnotte2013unidimensional}
Bonnotte, N. 2013.
\newblock \emph{Unidimensional and evolution methods for optimal
  transportation}.
\newblock Ph.D. thesis, Paris 11.

\bibitem[{Bousmalis et~al.(2016)Bousmalis, Trigeorgis, Silberman, Krishnan, and
  Erhan}]{bousmalis2016domain}
Bousmalis, K.; Trigeorgis, G.; Silberman, N.; Krishnan, D.; and Erhan, D. 2016.
\newblock Domain separation networks.
\newblock In \emph{Advances in Neural Information Processing Systems},
  343--351.

\bibitem[{Chen et~al.(2019)Chen, Chen, Jiang, and Jin}]{chen2019joint}
Chen, C.; Chen, Z.; Jiang, B.; and Jin, X. 2019.
\newblock Joint domain alignment and discriminative feature learning for
  unsupervised deep domain adaptation.
\newblock In \emph{Proc. of the AAAI Conf. on Artificial Intelligence},
  volume~33, 3296--3303.

\bibitem[{Chen and Liu(2018)}]{chen2018lifelong}
Chen, Z.; and Liu, B. 2018.
\newblock Lifelong machine learning.
\newblock \emph{Synthesis Lectures on Artificial Intelligence and Machine
  Learning}, 12(3): 1--207.

\bibitem[{Courty et~al.(2016)Courty, Flamary, Tuia, and
  Rakotomamonjy}]{courty2017optimal}
Courty, N.; Flamary, R.; Tuia, D.; and Rakotomamonjy, A. 2016.
\newblock Optimal transport for domain adaptation.
\newblock \emph{IEEE Transactions on Pattern Analysis and Machine
  Intelligence}, 39(9): 1853--1865.

\bibitem[{Daum{\'e}~III(2007)}]{daume2009frustratingly}
Daum{\'e}~III, H. 2007.
\newblock Frustratingly Easy Domain Adaptation.
\newblock In \emph{Proceedings of the 45th Annual Meeting of the Association of
  Computational Linguistics}, 256--263.

\bibitem[{Dong et~al.(2021)Dong, Cong, Sun, Fang, and Ding}]{dong2021and}
Dong, J.; Cong, Y.; Sun, G.; Fang, Z.; and Ding, Z. 2021.
\newblock Where and how to transfer: knowledge aggregation-induced
  transferability perception for unsupervised domain adaptation.
\newblock \emph{IEEE Transactions on Pattern Analysis and Machine
  Intelligence}.

\bibitem[{Dong et~al.(2020)Dong, Cong, Sun, Zhong, and Xu}]{dong2020can}
Dong, J.; Cong, Y.; Sun, G.; Zhong, B.; and Xu, X. 2020.
\newblock What can be transferred: Unsupervised domain adaptation for
  endoscopic lesions segmentation.
\newblock In \emph{Proceedings of the IEEE/CVF conference on computer vision
  and pattern recognition}, 4023--4032.

\bibitem[{Dredze and Crammer(2008)}]{dredze2008online}
Dredze, M.; and Crammer, K. 2008.
\newblock Online methods for multi-domain learning and adaptation.
\newblock In \emph{Proceedings of the Conference on Empirical Methods in
  Natural Language Processing}, 689--697. Association for Computational
  Linguistics.

\bibitem[{Ganin and Lempitsky(2015)}]{ganin2014unsupervised}
Ganin, Y.; and Lempitsky, V. 2015.
\newblock Unsupervised domain adaptation by backpropagation.
\newblock In \emph{Proceedings of ICML}, 1180--1189.

\bibitem[{Ganin et~al.(2016)Ganin, Ustinova, Ajakan, Germain, Larochelle,
  Laviolette, Marchand, and Lempitsky}]{ganin2016domain}
Ganin, Y.; Ustinova, E.; Ajakan, H.; Germain, P.; Larochelle, H.; Laviolette,
  F.; Marchand, M.; and Lempitsky, V. 2016.
\newblock Domain-adversarial training of neural networks.
\newblock \emph{The Journal of Machine Learning Research}, 17(1): 2096--2030.

\bibitem[{Ghifary et~al.(2016)Ghifary, Kleijn, Zhang, Balduzzi, and
  Li}]{ghifary2016deep}
Ghifary, M.; Kleijn, W.~B.; Zhang, M.; Balduzzi, D.; and Li, W. 2016.
\newblock Deep reconstruction-classification networks for unsupervised domain
  adaptation.
\newblock In \emph{European Conference on Computer Vision}, 597--613. Springer.

\bibitem[{Glorot, Bordes, and Bengio(2011)}]{glorot2011a}
Glorot, X.; Bordes, A.; and Bengio, Y. 2011.
\newblock Domain adaptation for large-scale sentiment classification: a deep
  learning approach.
\newblock In \emph{Proceedings of the 28th International Conference on Machine
  Learning}, 513--520.

\bibitem[{He et~al.(2016)He, Zhang, Ren, and Sun}]{he2016deep}
He, K.; Zhang, X.; Ren, S.; and Sun, J. 2016.
\newblock Deep residual learning for image recognition.
\newblock In \emph{Proceedings of the IEEE Conference on Computer Vision and
  Pattern Recognition}, 770--778.

\bibitem[{Jain and Learned-Miller(2011)}]{jain2011online}
Jain, V.; and Learned-Miller, E. 2011.
\newblock Online domain adaptation of a pre-trained cascade of classifiers.
\newblock In \emph{Proceedings of CVPR}, 577--584.

\bibitem[{Kang et~al.(2019)Kang, Jiang, Yang, and
  Hauptmann}]{kang2019contrastive}
Kang, G.; Jiang, L.; Yang, Y.; and Hauptmann, A.~G. 2019.
\newblock Contrastive adaptation network for unsupervised domain adaptation.
\newblock In \emph{Proceedings of the IEEE Conference on Computer Vision and
  Pattern Recognition}, 4893--4902.

\bibitem[{Kirkpatrick et~al.(2017)Kirkpatrick, Pascanu, Rabinowitz, Veness,
  Desjardins, Rusu, Milan, Quan, Ramalho, Grabska-Barwinska
  et~al.}]{kirkpatrick2017overcoming}
Kirkpatrick, J.; Pascanu, R.; Rabinowitz, N.; Veness, J.; Desjardins, G.; Rusu,
  A.~A.; Milan, K.; Quan, J.; Ramalho, T.; Grabska-Barwinska, A.; et~al. 2017.
\newblock Overcoming catastrophic forgetting in neural networks.
\newblock \emph{Proceedings of the national academy of sciences}, 114(13):
  3521--3526.

\bibitem[{Kolouri, Rohde, and Hoffman(2018)}]{kolouri2018sliced}
Kolouri, S.; Rohde, G.~K.; and Hoffman, H. 2018.
\newblock Sliced {W}asserstein Distance for Learning {G}aussian Mixture Models.
\newblock In \emph{IEEE Conference on Computer Vision and Pattern Recognition},
  3427--.

\bibitem[{Kundu et~al.(2020)Kundu, Venkat, Babu et~al.}]{kundu2020universal}
Kundu, J.~N.; Venkat, N.; Babu, R.~V.; et~al. 2020.
\newblock Universal Source-Free Domain Adaptation.
\newblock In \emph{Proceedings of the IEEE/CVF Conference on Computer Vision
  and Pattern Recognition}, 4544--4553.

\bibitem[{Lee et~al.(2019)Lee, Batra, Baig, and Ulbricht}]{lee2019sliced}
Lee, C.-Y.; Batra, T.; Baig, M.~H.; and Ulbricht, D. 2019.
\newblock Sliced wasserstein discrepancy for unsupervised domain adaptation.
\newblock In \emph{Proceedings of the IEEE Conference on Computer Vision and
  Pattern Recognition}, 10285--10295.

\bibitem[{Li et~al.(2020{\natexlab{a}})Li, Zhai, Luo, Ge, and
  Ren}]{li2020enhanced}
Li, M.; Zhai, Y.-M.; Luo, Y.-W.; Ge, P.-F.; and Ren, C.-X. 2020{\natexlab{a}}.
\newblock Enhanced transport distance for unsupervised domain adaptation.
\newblock In \emph{Proceedings of the IEEE/CVF Conf. on Computer Vision and
  Pattern Recognition}, 13936--13944.

\bibitem[{Li et~al.(2020{\natexlab{b}})Li, Jiao, Cao, Wong, and
  Wu}]{li2020model}
Li, R.; Jiao, Q.; Cao, W.; Wong, H.-S.; and Wu, S. 2020{\natexlab{b}}.
\newblock Model Adaptation: Unsupervised Domain Adaptation without Source Data.
\newblock In \emph{Proceedings of the IEEE/CVF Conference on Computer Vision
  and Pattern Recognition}, 9641--9650.

\bibitem[{Liang, Hu, and Feng(2020)}]{liang2020we}
Liang, J.; Hu, D.; and Feng, J. 2020.
\newblock Do we really need to access the source data? source hypothesis
  transfer for unsupervised domain adaptation.
\newblock In \emph{International Conference on Machine Learning}, 6028--6039.
  PMLR.

\bibitem[{Liu, Zhang, and Wang(2021)}]{liu2021source}
Liu, Y.; Zhang, W.; and Wang, J. 2021.
\newblock Source-free domain adaptation for semantic segmentation.
\newblock In \emph{Proceedings of the IEEE/CVF Conference on Computer Vision
  and Pattern Recognition}, 1215--1224.

\bibitem[{Long et~al.(2015)Long, Cao, Wang, and Jordan}]{long2015learning}
Long, M.; Cao, Y.; Wang, J.; and Jordan, M. 2015.
\newblock Learning Transferable Features with Deep Adaptation Networks.
\newblock In \emph{Proceedings of International Conference on Machine
  Learning}, 97--105.

\bibitem[{Long et~al.(2018)Long, Cao, Wang, and Jordan}]{long2018conditional}
Long, M.; Cao, Z.; Wang, J.; and Jordan, M.~I. 2018.
\newblock Conditional adversarial domain adaptation.
\newblock In \emph{Advances in Neural Information Processing Systems},
  1640--1650.

\bibitem[{Long et~al.(2017)Long, Zhu, Wang, and Jordan}]{long2017deep}
Long, M.; Zhu, H.; Wang, J.; and Jordan, M.~I. 2017.
\newblock Deep transfer learning with joint adaptation networks.
\newblock In \emph{Proceedings of the 34th International Conference on Machine
  Learning-Volume 70}, 2208--2217. JMLR. org.

\bibitem[{Ma, Zhang, and Xu(2019)}]{ma2019deep}
Ma, X.; Zhang, T.; and Xu, C. 2019.
\newblock Deep multi-modality adversarial networks for unsupervised domain
  adaptation.
\newblock \emph{IEEE Transactions on Multimedia}, 21(9): 2419--2431.

\bibitem[{McInnes et~al.(2018)McInnes, Healy, Saul, and
  Gro{\ss}berger}]{mcinnes2018umap}
McInnes, L.; Healy, J.; Saul, N.; and Gro{\ss}berger, L. 2018.
\newblock {UMAP}: Uniform Manifold Approximation and Projection.
\newblock \emph{Journal of Open Source Soft.}, 3(29): 861.

\bibitem[{Mirza et~al.(2022)Mirza, Micorek, Possegger, and
  Bischof}]{mirza2022norm}
Mirza, M.~J.; Micorek, J.; Possegger, H.; and Bischof, H. 2022.
\newblock The Norm Must Go On: Dynamic Unsupervised Domain Adaptation by
  Normalization.
\newblock In \emph{Proceedings of the IEEE/CVF Conference on Computer Vision
  and Pattern Recognition}, 14765--14775.

\bibitem[{Neal(2003)}]{neal2003slice}
Neal, R. 2003.
\newblock Slice sampling.
\newblock \emph{Annals of statistics}, 705--741.

\bibitem[{Oliveira, Minku, and Oliveira(2019)}]{oliveira2019gmm}
Oliveira, G.~H.; Minku, L.~L.; and Oliveira, A.~L. 2019.
\newblock GMM-VRD: A Gaussian Mixture model for dealing with virtual and real
  concept drifts.
\newblock In \emph{2019 International Joint Conference on Neural Networks
  (IJCNN)}, 1--8. IEEE.

\bibitem[{Pei et~al.(2018)Pei, Cao, Long, and Wang}]{pei2018multi}
Pei, Z.; Cao, Z.; Long, M.; and Wang, J. 2018.
\newblock Multi-adversarial domain adaptation.
\newblock In \emph{Proceedings Thirty-Second AAAI Conference on Artificial
  Intelligence}, 3934--3941.

\bibitem[{Pf{\"u}lb and Gepperth(2021)}]{pfulb2021overcoming}
Pf{\"u}lb, B.; and Gepperth, A. 2021.
\newblock Overcoming catastrophic forgetting with gaussian mixture replay.
\newblock In \emph{2021 International Joint Conference on Neural Networks
  (IJCNN)}, 1--9. IEEE.

\bibitem[{Redko and Sebban(2017)}]{redko2017theoretical}
Redko, A., I.and~Habrard; and Sebban, M. 2017.
\newblock Theoretical analysis of domain adaptation with optimal transport.
\newblock In \emph{Joint European Conference on Machine Learning and Knowledge
  Discovery in Databases}, 737--753. Springer.

\bibitem[{Rostami(2021)}]{rostami2021lifelong}
Rostami, M. 2021.
\newblock Lifelong domain adaptation via consolidated internal distribution.
\newblock In \emph{Advances in Neural Information Processing Systems},
  volume~34, 11172--11183.

\bibitem[{Rostami, Isele, and Eaton(2020)}]{rostami2020using}
Rostami, M.; Isele, D.; and Eaton, E. 2020.
\newblock Using task descriptions in lifelong machine learning for improved
  performance and zero-shot transfer.
\newblock \emph{Journal of Artificial Intelligence Research}, 67: 673--704.

\bibitem[{Rostami et~al.(2019)Rostami, Kolouri, Eaton, and
  Kim}]{rostami2019deep}
Rostami, M.; Kolouri, S.; Eaton, E.; and Kim, K. 2019.
\newblock Deep transfer learning for few-shot sar image classification.
\newblock \emph{Remote Sensing}, 11(11): 1374.

\bibitem[{Rostami et~al.(2018)Rostami, Kolouri, Kim, and
  Eaton}]{rostami2018multi}
Rostami, M.; Kolouri, S.; Kim, K.; and Eaton, E. 2018.
\newblock Multi-Agent Distributed Lifelong Learning for Collective Knowledge
  Acquisition.
\newblock In \emph{Proceedings of the 17th International Conference on
  Autonomous Agents and MultiAgent Systems}, 712--720.

\bibitem[{Rostami et~al.(2020)Rostami, Kolouri, Pilly, and
  McClelland}]{rostami2020generative}
Rostami, M.; Kolouri, S.; Pilly, P.; and McClelland, J. 2020.
\newblock Generative continual concept learning.
\newblock In \emph{Proceedings of the AAAI Conference on Artificial
  Intelligence}, volume~34, 5545--5552.

\bibitem[{Saito et~al.(2020)Saito, Kim, Sclaroff, and
  Saenko}]{saito2020universal}
Saito, K.; Kim, D.; Sclaroff, S.; and Saenko, K. 2020.
\newblock Universal domain adaptation through self supervision.
\newblock \emph{Advances in neural information processing systems}, 33:
  16282--16292.

\bibitem[{Sankaranarayanan et~al.(2018)Sankaranarayanan, Balaji, Castillo, and
  Chellappa}]{sankaranarayanan2018generate}
Sankaranarayanan, S.; Balaji, Y.; Castillo, C.~D.; and Chellappa, R. 2018.
\newblock Generate to adapt: Aligning domains using generative adversarial
  networks.
\newblock In \emph{Proceedings of CVPR}, 8503--8512.

\bibitem[{Shen et~al.(2018)Shen, Qu, Zhang, and Yu}]{shen2018wasserstein}
Shen, J.; Qu, Y.; Zhang, W.; and Yu, Y. 2018.
\newblock Wasserstein distance guided representation learning for domain
  adaptation.
\newblock In \emph{Proceedings of the AAAI Conference on Artificial
  Intelligence}, volume~32.

\bibitem[{Stan and Rostami(2021)}]{stan2021unsupervised}
Stan, S.; and Rostami, M. 2021.
\newblock Unsupervised model adaptation for continual semantic segmentation.
\newblock In \emph{Proceedings of the AAAI Conference on Artificial
  Intelligence}, volume~35, 2593--2601.

\bibitem[{Stan and Rostami(2022{\natexlab{a}})}]{stan2021privacy}
Stan, S.; and Rostami, M. 2022{\natexlab{a}}.
\newblock Privacy preserving domain adaptation for semantic segmentation of
  medical images.

\bibitem[{Stan and Rostami(2022{\natexlab{b}})}]{stan2021secure}
Stan, S.; and Rostami, M. 2022{\natexlab{b}}.
\newblock Secure Domain Adaptation with Multiple Sources.
\newblock \emph{Transactions on Machine Learning Research}.

\bibitem[{Sun, Feng, and Saenko(2016)}]{sun2016return}
Sun, B.; Feng, J.; and Saenko, K. 2016.
\newblock Return of frustratingly easy domain adaptation.
\newblock In \emph{Proceedings of the AAAI Conf. on Artificial Intelligence},
  volume~30.

\bibitem[{Sun and Saenko(2016)}]{sun2016deep}
Sun, B.; and Saenko, K. 2016.
\newblock Deep coral: Correlation alignment for deep domain adaptation.
\newblock In \emph{European conference on computer vision}, 443--450. Springer.

\bibitem[{Tzeng et~al.(2017)Tzeng, Hoffman, Saenko, and
  Darrell}]{tzeng2017adversarial}
Tzeng, E.; Hoffman, J.; Saenko, K.; and Darrell, T. 2017.
\newblock Adversarial discriminative domain adaptation.
\newblock In \emph{Proceedings of the IEEE Conference on Computer Vision and
  Pattern Recognition}, 7167--7176.

\bibitem[{Vorburger and Bernstein(2006)}]{vorburger2006entropy}
Vorburger, P.; and Bernstein, A. 2006.
\newblock Entropy-based concept shift detection.
\newblock In \emph{Sixth International Conference on Data Mining (ICDM'06)},
  1113--1118. IEEE.

\bibitem[{Wang et~al.(2022)Wang, Wu, Weng, Chen, Qi, and Jiang}]{wang2022cross}
Wang, R.; Wu, Z.; Weng, Z.; Chen, J.; Qi, G.-J.; and Jiang, Y.-G. 2022.
\newblock Cross-domain contrastive learning for unsupervised domain adaptation.
\newblock \emph{IEEE Transactions on Multimedia}.

\bibitem[{Wang and Zhang(2020)}]{wang2020self}
Wang, S.; and Zhang, L. 2020.
\newblock Self-adaptive re-weighted adversarial domain adaptation.
\newblock \emph{arXiv preprint arXiv:2006.00223}.

\bibitem[{Wu(2016)}]{wu2016online}
Wu, D. 2016.
\newblock Online and offline domain adaptation for reducing BCI calibration
  effort.
\newblock \emph{IEEE Transactions on Human-Machine Systems}, 47(4): 550--563.

\bibitem[{Wu, Hua, and Zhang(2005)}]{wu2005tracking}
Wu, J.; Hua, X.-S.; and Zhang, B. 2005.
\newblock Tracking concept drifting with Gaussian mixture model.
\newblock In \emph{Visual Communications and Image Processing 2005}, volume
  5960, 1562--1570. SPIE.

\bibitem[{Zhang et~al.(2019)Zhang, Tang, Jia, and Tan}]{zhang2019domain}
Zhang, Y.; Tang, H.; Jia, K.; and Tan, M. 2019.
\newblock Domain-symmetric networks for adversarial domain adaptation.
\newblock In \emph{Proceedings of the IEEE Conference on Computer Vision and
  Pattern Recognition}, 5031--5040.

\end{thebibliography}

 \clearpage

 \appendix

\title{S}

\section{Technical Appendix}

\subsection{Sliced Wasserstein distance}

 SWD is inspired by the definition of the  Wasserstein distance (WD) or the optimal transport metric.  The optimal transport metric between two probability distributions  $p_{\mathcal{S}}$ and $p_{\mathcal{T}}$ is defined as:
\begin{equation}
W_c(p_{\mathcal{S}},p_{\mathcal{T}})=\text{inf}_{\gamma\in \Gamma(p_{\mathcal{S}},p_{\mathcal{T}})} \int_{{X}\times {Y}} c(x,y)d\gamma(x,y)
\label{eq:kantorovich}
\end{equation}
where $\Gamma(p_{\mathcal{S}},p_{\mathcal{T}})$ is the set of all joint distributions $p_{{\mathcal{S}},{\mathcal{T}}}$  with marginal single variable distributions $p_{\mathcal{S}}$ and $p_{\mathcal{T}}$, and $c:X\times Y\rightarrow \mathbb{R}^+$ is the transportation cost which normally is assumed to be $\ell_2$-norm Euclidean distance.
Computing the integral in Eq.~\ref{eq:kantorovich} requires solving an optimization problem which is a linear programming problem. 
However, when the distributions are 
 $1-$dimensional,  Eq.~\ref{eq:kantorovich} has closed-form solution:
\begin{equation}
W_c(p_{\mathcal{S}},p_{\mathcal{T}})= \int_{0}^1 c(P_{\mathcal{S}}^{-1}(\tau),P_{\mathcal{T}}^{-1}(\tau))d\tau,
\label{eq:oneD}
\end{equation} 
where $P_{\mathcal{S}}$ and $P_{\mathcal{T}}$ are the cumulative distributions of   distributions $p_{\mathcal{S}}$ and $p_{\mathcal{T}}$. 
This closed-form solution  motivates the definition of SWD in order to reduce computations when we have higher dimensional distributions. 

The idea behind the SWD is based on the slice sampling~\cite{neal2003slice}. The idea is  to project two $d$-dimensional probability distributions into their marginal one-dimensional distributions to generate $1-$dimensional  marginal probability distributions so we can benefit from the closed form solution. 
For the distribution $p_\mathcal{S}$,  a one-dimensional slice of the distribution is defined:
\begin{equation}
\mathcal{R}p_\mathcal{S}(t;\bm{\gamma})=\int_{\mathcal{S}^{d-1}} p_\mathcal{S}(\bm{x})\bm{\delta}(t-\langle\bm{\gamma}, \bm{x}\rangle)d\bm{x},
\label{eq:radon}
\end{equation}
where $\bm{\delta}(\cdot)$ denotes the Kronecker delta function,  $\langle \cdot ,\cdot\rangle$ denotes the vector inner dot product, $\mathbb{S}^{d-1}$ is the $d$-dimensional unit sphere, and $\bm{\gamma}$ is the projection direction. In other words, $\mathcal{R}p_\mathcal{S}(\cdot;\bm{\gamma)}$ is a marginal distribution of $p_\mathcal{S}$  obtained from integrating $p_\mathcal{S}$ over the hyperplanes orthogonal to $\bm{\gamma}$. The SWD then is defined as integral of the  sliced distributions  over all $1-$dimensional subspaces $\bm{\gamma}$ on the unit spehere:
\begin{eqnarray}
SW(p_\mathcal{S},p_\mathcal{T})=   \int_{\mathbb{S}^{d-1}} W(\mathcal{R} p_\mathcal{S}(\cdot;\gamma),\mathcal{R} p_\mathcal{T}(\cdot;\gamma))d\gamma
\label{eq:radonSWDdistance}
\end{eqnarray}
 where $W(\cdot)$ denotes the Wasserstein distance.
The main advantage of using the SWD is that  calculation of the SWD does not require solving a numerically expensive optimization.   Since only samples from distributions are available, the  one-dimensional Wasserstein distance can be approximated as the $\ell_p$-distance between the sorted samples. 
Note however, this way we can compute merely the integrand function in Eq.~\eqref{eq:radonSWDdistance} for a known $\gamma$.
To approximate the integral in Eq.~\eqref{eq:radonSWDdistance}, we can  use a Monte Carlo style integration. First, we sample the projection subspace $\bm{\gamma}$ from a uniform distribution that is defined over the unit sphere and then compute $1-$dimensional Wasserstein distance on the sample. We can then approximate the integral in Eq.~\eqref{eq:radonSWDdistance}  by computing the arithmetic average over a suitably large enough  number of drawn samples.   Formally, the SWD between $f$-dimensional samples $\{\phi(\bm{x}_i^\mathcal{S})\in \mathbb{R}^f\sim p_\mathcal{S}\}_{i=1}^M$ and $\{\phi(\bm{x}_i^\mathcal{T})\in \mathbb{R}^f \sim p_\mathcal{T}\}_{j=1}^M$ in our problem of interest can be approximated as the following sum:
\begin{equation}
SW^2(p_\mathcal{S},p_\mathcal{T})\approx \frac{1}{L}\sum_{l=1}^L \sum_{i=1}^M| \langle\gamma_l, \phi(\bm{x}_{s_l[i]}^\mathcal{S}\rangle)- \langle\gamma_l, \phi(\bm{x}_{t_l[i]}^\mathcal{T})\rangle|^2
\label{eq:SWDempirical}
\end{equation}
where $\gamma_l\in\mathbb{S}^{f-1}$ is uniformly drawn random sample from the unit $f$-dimensional ball $\mathbb{S}^{f-1}$, and  $s_l[i]$ and $t_l[i]$ are the sorted indices of $\{\gamma_l\cdot\phi(\bm{x}_i)\}_{i=1}^M$ for source and target domains, respectively. We utilize the empirical version of SWD in Eq.~\eqref{eq:SWDempirical} as the discrepancy measure between the   probability distributions to match them in the embedding space. Note that   the function in Eq.~\eqref{eq:SWDempirical} is differentiable with respect to the encoder parameters and hence we can use gradient-based optimization techniques that are commonly used in deep learning to minimize it with respect to the model parameters.

 \section{Proof of Theorem~1}

We first note that When we generate the pseudo-dataset, we ensure to select  pseudo-data points for which the model is confident. To this end, we pick a threshold $\tau$, draw random pseudo-data points $\bm{z}_i^{p}$, and pass them through the classifier subnetwork. We then look at the predicted label distribution at the final softmax layer and include only those data-points for which the model is confident with prediction probability greater than $\tau$. Let $e_{\mathcal{P}}$ denotes the true risk of the initial optimal model that is trained using the source domain data on the generated pseudo-dataset.  

We benefit from  the following theorem by Redko et al.~\cite{redko2017theoretical} in our proof.  

\textbf{Theorem~2 (Redko et al.~\cite{redko2017theoretical})}: Under the assumptions described in our framework, assume that a model is trained on the source domain, then for any $d'>d$ and $\zeta<\sqrt{2}$, there exists a constant number $N_0$ depending on $d'$ such that for any  $\bm{x}i>0$ and $\min(N,M)\ge \max (\bm{x}i^{-(d'+2),1})$ with probability at least $1-\bm{x}i$, the following holds:
\begin{equation}
\begin{split}
e_{\mathcal{T}}\le & e_{\mathcal{S}} +W(\hat{\mu}_{\mathcal{T}},\hat{\mu}_{\mathcal{S}})+e_{\mathcal{C}}(\bm{w}^*)+ \\& \sqrt{\big(2\log(\frac{1}{\bm{x}i})/\zeta\big)}\big(\sqrt{\frac{1}{N}}+\sqrt{\frac{1}{M}}\big).
\end{split}
\label{eq:theroemfromcourty}
\end{equation}    
 
Theorem~2 provides an upperbound for the performance of the source domain trained model on the target domain. We use Theorem~2 to deduce Theorem~1.
 Redko et al.~\cite{redko2017theoretical} prove  Theorem~2 for a binary classification setting in a joint training UDA setting. We also provide  our proof in this case but it can be conveniently extended.

\textbf{Theorem~1 }:  Consider that we generate a pseudo-dataset  using the internal distribution and update the model for sequential model adaptation using SMAUP algorithm. Then, the following   holds:
\begin{equation}
\small
\begin{split}
e_{\mathcal{T}}\le & e_{\mathcal{S}} +W(\hat{\mu}_{\mathcal{S}},\hat{\mu}_{\mathcal{P}})+W(\hat{\mu}_{\mathcal{T}},\hat{\mu}_{\mathcal{P}})+(1-\tau)+e_{\mathcal{C'}}(\bm{w}^*)+\\&\sqrt{\big(2\log(\frac{1}{\bm{x}i})/\zeta\big)}\big(\sqrt{\frac{1}{N}}+\sqrt{\frac{1}{M }}+2\sqrt{\frac{1}{N_p }}\big),
\end{split}
\label{eq:theroemforPLnipsApp}
\end{equation}    
where    $\bm{x}i$ is a constant which depends on $\mathcal{L}(\cdot)$ and  $e_{C'}(\bm{w}^*)$  denotes the expected risk of the optimally joint trained model when used   on both the source domain and  the    pseudo-dataset.

\textbf{Proof:}   Since the parameter $\tau$ denotes the threshold that we use to select the pseudo-data points in the embedding space,  then the probability that the  predicted  labels for the pseudo-data points to be false by this model is equal to $1-\tau$.  We can define the following difference for the pseudo-data points:
\begin{equation}
\begin{split}
  |\mathcal{L}(h_{\bm{w}_0}(\bm{z}^p_i),\bm{y}^p_i)- \mathcal{L}(h_{\bm{w}_0}(\bm{z}^p_i),\hat{\bm{y}}_i^{p})|= \begin{cases}
    0, & \text{if $\bm{y}^t_i=\hat{\bm{y}}_i^{t}$}.\\
    1, & \text{otherwise}.
  \end{cases}
\end{split}
\label{eq:theroemforPLproof}
\end{equation}    
 Hence, using Jensen's inequality and by applying the expectation operator on the above error can be computed as:
\begin{equation}
\begin{split}
|e_{\mathcal{P}}-e_{\mathcal{T}}|\le\mathbb{E}\big(|\mathcal{L}(h_{\bm{w}_0}(\bm{z}^p_i),\bm{y}^p_i)-& \mathcal{L}(h_{\bm{w}_0}(\bm{z}^p_i),\hat{\bm{y}}_i^{p})|\big) \\&
\le(1-\tau).
\end{split}
\label{eq:theroemforPLproofexpectation}
\end{equation}    
Using Eq.~\eqref{eq:theroemforPLproofexpectation} we can deduce:
\begin{equation}
\begin{split}
&e_{\mathcal{S}}+e_{\mathcal{T}}=e_{\mathcal{S}}+e_{\mathcal{T}}+e_{\mathcal{P}}-e_{\mathcal{P}}\le  
e_{\mathcal{S}}+e_{\mathcal{P}}+|e_{\mathcal{T}}-e_{\mathcal{P}}|\le\\&  
e_{\mathcal{S}}+e_{\mathcal{P}}+(1-\tau).
\end{split}
\label{eq:theroemforPLprooftrangleinq}
\end{equation}    
Note that since Eq.~\eqref{eq:theroemforPLprooftrangleinq} is valid for all $\bm{w}$, if we consider the joint optimal parameter $\bm{w}^*$ in the right-hand and the left-hand sides of Eq.~\eqref{eq:theroemforPLprooftrangleinq}, we deduce:
\begin{equation}
\begin{split}
e_C(\bm{w}^*)\le e_{C'}(\bm{w})+(1-\tau).
\end{split}
\label{eq:theroemforPLprooftartplerror}
\end{equation}    
Now by considering
Theorem~2 for the t he source and the target domains and then applying Eq.~\eqref{eq:theroemforPLprooftartplerror}  on Eq.\eqref{eq:theroemfromcourty},  we have:
\begin{equation}
\begin{split}
e_{\mathcal{T}}\le & e_{\mathcal{S}} +W(\hat{\mu}_{\mathcal{T}},\hat{\mu}_{\mathcal{S}})+e_{\mathcal{C}}'(\bm{w}^*)+ (1-\tau) \\&+ \sqrt{\big(2\log(\frac{1}{\bm{x}i})/\zeta\big)}\big(\sqrt{\frac{1}{N}}+\sqrt{\frac{1}{M}}\big).
\end{split}
\label{eq:theroemfromcourty1}
\end{equation}    
We used Eq.~\eqref{eq:theroemfromcourty1} to deduce Theorem~1 by relating the terms to the internal distribution.

We first  use the triangular inequality on the WD metric to deduce the following relation for the $W(\hat{\mu}_{\mathcal{T}},\hat{\mu}_{\mathcal{S}})$ term in Eq.~\eqref{eq:theroemfromcourty1}:
\begin{equation}
\begin{split}
& W(\hat{\mu}_{\mathcal{T}},\hat{\mu}_{\mathcal{S}})\le  W(\hat{\mu}_{\mathcal{T}},\mu_{\mathcal{P}})+W(\hat{\mu}_{\mathcal{S}},\mu_{\mathcal{P}})  \le\\& W(\hat{\mu}_{\mathcal{T}},\hat{\mu}_{\mathcal{P}})+W(\hat{\mu}_{\mathcal{S}},\hat{\mu}_{\mathcal{P}})+2W(\hat{\mu}_{\mathcal{P}},\mu_{\mathcal{P}}) .
\end{split}
\label{eq:theroemfromcourty2}
\end{equation}

  We can   simplify the term $W(\hat{\mu}_{\mathcal{P}},\mu_{\mathcal{P}})$ in the above using Theorem 1.1 in the work by Bolley et al.~\cite{bolley2007quantitative}.

  \textbf{Theorem~3} (Theorem 1.1 by Bolley et al.~\cite{bolley2007quantitative}): consider that $p(\cdot) \in\mathcal{P}(\mathcal{Z})$ and $\int_{\mathcal{Z}} \exp{(\alpha \|\bm{x}\|^2_2)}dp(\bm{x})<\infty$ for some $\alpha>0$. Let $\hat{p}(\bm{x})=\frac{1}{N}\sum_i\delta(\bm{x}_i)$ denote the empirical distribution that is built from the samples $\{\bm{x}_i\}_{i=1}^N$ that are drawn i.i.d from $\bm{x}_i\sim p(\bm{x})$. Then for any $d'>d$ and $\bm{x}i<\sqrt{2}$, there exists $N_0$ such that for any $\epsilon>0$ and $N\ge N_o\max(1,\epsilon^{-(d'+2)})$, we have:
 \begin{equation}
\begin{split}
P(W(p,\hat{p})>\epsilon)\le \exp(-\frac{-\bm{x}i}{2}N\epsilon^2)
\end{split}
\label{eq:mainSuppICML3}
\end{equation}  
 This relation measure the distance between the estimated empirical distribution and the true distribution in terms of the WD distance. We can use it estimate the error between the true and the empirical distribution in terms of the distribution samples used for empirical estimation to simplify $W(\hat{\mu}_{\mathcal{P}},\mu_{\mathcal{P}})$ in Eq.~\eqref{eq:theroemfromcourty1}.

 Finally, replacing  Eq.~\eqref{eq:theroemfromcourty2} and Eq.~\eqref{eq:mainSuppICML3} in the corresponding terms Eq.~\eqref{eq:theroemfromcourty1},     concludes Theorem~2 as stated:
\begin{equation}
\small
\begin{split}
e_{\mathcal{T}}\le & e_{\mathcal{S}} +W(\hat{\mu}_{\mathcal{S}},\hat{\mu}_{\mathcal{P}})+W(\hat{\mu}_{\mathcal{T}},\hat{\mu}_{\mathcal{P}})+(1-\tau)+e_{\mathcal{C'}}(\bm{w}^*)\\&+\sqrt{\big(2\log(\frac{1}{\bm{x}i})/\zeta\big)}\big(\sqrt{\frac{1}{N}}+\sqrt{\frac{1}{M }}+2\sqrt{\frac{1}{N_p }}\big),
\end{split}
\label{eq:theroemforPLnips55}
\end{equation}

\subsection{Details of Experimental Implementation }

We validate our method on five  standard benchmark set of UDA tasks and adapt them for sequential task learning. 
 
\textbf{Digit recognition   tasks:}  MNIST ($\mathcal{M}$),
USPS ($\mathcal{U}$), and
Street View House Numbers, i.e., SVHN ($\mathcal{S}$), datasets are used as the three digit recognition  domains. Following the  the UDA literature, we report performance on the three UDA tasks: $\mathcal{M}\rightarrow \mathcal{U}$, $\mathcal{U}\rightarrow \mathcal{M}$, and $\mathcal{S}\rightarrow \mathcal{M}$ tasks. 

\textbf{Office-31 Detest:} this dataset   is a visual recognition dataset with a total of  $4,652$ images which are categorized into ten classes in three distinct domains: Amazon ($\mathcal{A}$), Webcam ($\mathcal{W}$) and DSLR ($\mathcal{D}$). We report performance on  the six pair-wise definable UDA tasks among these domains.

\textbf{ImageCLEF-DA Dataset:} this image classification dataset is a generated using   the 12 shared classes between the Caltech-256 ($\mathcal{C}$), the   ILSVRC 2012 ($\mathcal{I}$), and the Pascal VOC 2012 ($\mathcal{P}$) visual recognition datasets. This dataset is fully balanced in which each class has 50 images which results in 600 images for each domian.   We perform experiments on the six possible UDA tasks.

\textbf{Office-Caltech Dataset}: this dataset is generated using the 10 shared classes between the Office-31 and Caltech-256 datasets.  There are four domains A, C, D, W with 12 definable tasks and    2533  images in total. These tasks help to measure the cross-dataset adaptation performance.

\textbf{VisDA-2017:}  the goal for this dataset is to train a model on a synthetic domain and adapt it to work well on real natural images.  The synthetic images are generated by renderings of 3D models from different angles and  lightning conditions across 12 classes with over 280K images.

In the digit recognition experiments, we resized the images of SVHN dataset to $28\times 28$ images to have the same size of the MNIST and the USPS datasets. This is necessary because we use the same encoder across all domains.

In our experiments, we used cross entropy loss as the discrimination loss.  At each   training epoch, we computed the combined loss function on the training split of data and stopped training when the loss function became constant.  We used Keras for implementation and   ADAM optimizer with learning rate $lr=10^{-4}$.  We have run our code on a cluster node equipped with 2 Nvidia Tesla P100-SXM2 GPU's.  
We used the  classification rate on  the  testing set to measure performance of the algorithms.   We performed 10 training trials   and reported the average performance and the standard deviation.

\end{document}